\documentclass[journal]{IEEEtran}
\usepackage{xcolor,soul,framed} 
\usepackage{booktabs}
\colorlet{shadecolor}{yellow}
\usepackage[pdftex]{graphicx}
\graphicspath{{../pdf/}{../jpeg/}}
\DeclareGraphicsExtensions{.pdf,.jpeg,.png}

\usepackage[cmex10]{amsmath}
\usepackage{cite}
\usepackage{array}
\usepackage{comment}
\usepackage{mdwmath}
\usepackage{mdwtab}
\usepackage{eqparbox}
\usepackage{url}
\usepackage{lscape}
\usepackage{multirow}
\usepackage{tikz}
\newcommand{\tabitem}{~~\llap{\textbullet}~~}
\usepackage{amssymb}
\usepackage{pifont}
\usepackage{xcolor}
\usepackage{xcolor,pifont}
\usepackage{subcaption}
\usepackage{booktabs}

\usepackage{acronym}
\usepackage[none]{hyphenat}
\usepackage{balance}

\acrodef{ML}{Machine Learning}
\acrodef{DL}{Deep Learning}
\acrodef{AI}{Artificial Intelligence}
\acrodef{QoS}{Quality of Service}
\acrodef{IoT}{Internet of Things}
\acrodef{FL}{Federated Learning}
\acrodef{FML}{Federated Machine Learning}
\acrodef{IoMT}{Internet of Medical Things}
\acrodef{EHR}{Electronic Health Records}
\acrodef{NLP}{Natural Language Processing}
\acrodef{MEC}{Mobile Edge Computing}
\acrodef{EMR}{Electronic Medical Record}
\acrodef{EMRs}{Electronic Medical Records}
\acrodef{MRI}{Magnetic Resonance Imaging}
\acrodef{CNN}{Convolutional Neural Network}
\acrodef{GCAE}{Generative Convolutional Autoencoder}
\acrodef{GDPR}{General Data Protection Regulation}
\acrodef{HIPAA}{Health Insurance Portability and Accountability Act}
\acrodef{IPFS}{Interplanetary File System}
\acrodef{CT}{Computed Tomography}
\acrodef{TFF}{Tensor Flow Federated}
\acrodef{FMs}{Foundational Models}

\begin{document}
\bstctlcite{IEEEexample:BSTcontrol}
    \title{Federated Learning for Medical Applications: A Taxonomy, Current Trends, Challenges, and Future Research Directions}
  \author{Ashish Rauniyar,~\IEEEmembership{Member,~IEEE,} Desta Haileselassie Hagos,~\IEEEmembership{Member,~IEEE,} Debesh Jha,~\IEEEmembership{Member,~IEEE,} Jan Erik Håkegård,~\IEEEmembership{Senior Member,~IEEE,} Ulas Bagci,~\IEEEmembership{Senior Member,~IEEE,}  Danda B. Rawat,~\IEEEmembership{Senior Member,~IEEE,} Vladimir Vlassov,~\IEEEmembership{Senior Member,~IEEE}
  
\thanks{A. Rauniyar and J. E. Håkegård are with Sustainable Communication Technologies (SCT), SINTEF Digital, Trondheim, 7034, Norway (e-mail: ashish.rauniyar@sintef.no; jan.e.hakegard@sintef.no).} 
\thanks{D. H. Hagos and DB. Rawat are with the DoD Center of Excellence in Artificial Intelligence and Machine Learning (CoE-AIML), College of Engineering and Architecture (CEA), Department of Electrical Engineering and Computer Science, Howard University, Washington DC, USA (e-mail: desta.hagos@howard.edu; danda.rawat@howard.edu).}
\thanks{D. Jha and U. Bagci are with Machine \& Hybrid Intelligence Lab, Department of Radiology, Northwestern University, USA (e-mail: debesh.jha@northwestern.edu; ulas.bagci@northwestern.edu).}
\thanks{V. Vlassov is with the Department of Computer Science, School of Electrical Engineering and Computer Science, KTH Royal Institute of Technology, Stockholm, Sweden (e-mail: vladv@kth.se).}

\thanks{This research work was supported by the COPS (Comprehensive Privacy and Security for Resilient CPS/IoT) project funded by the Research Council of Norway under project number:300102.}
\thanks{Accepted for publication in IEEE Internet of Things Journal.}


}

\maketitle

\begin{abstract}
With the advent of the \ac{IoT}, \ac{AI}, \ac{ML}, and \ac{DL} algorithms, the landscape of data-driven medical applications has emerged as a promising avenue for designing robust and scalable diagnostic and prognostic models from medical data. This has gained a lot of attention from both academia and industry, leading to significant improvements in healthcare quality. However, the adoption of AI-driven medical applications still faces tough challenges, including meeting security, privacy, and quality of service (QoS) standards. Recent developments in \ac{FL} have made it possible to train complex machine-learned models in a distributed manner and has become an active research domain, particularly processing the medical data at the edge of the network in a decentralized way to preserve privacy and address security concerns. To this end, in this paper, we explore the present and future of FL technology in medical applications where data sharing is a significant challenge.  We delve into the current research trends and their outcomes, unravelling the complexities of designing reliable and scalable \ac{FL} models. Our paper outlines the fundamental statistical issues in FL, tackles device-related problems, addresses security challenges, and navigates the complexity of privacy concerns, all while highlighting its transformative potential in the medical field. Our study primarily focuses on medical applications of \ac{FL}, particularly in the context of global cancer diagnosis. We highlight the potential of FL to enable computer-aided diagnosis tools that address this challenge with greater effectiveness than traditional data-driven methods. Recent literature has shown that FL models are robust and generalize well to new data, which is essential for medical applications. We hope that this comprehensive review will serve as a checkpoint for the field, summarizing the current state-of-the-art and identifying open problems and future research directions.

\end{abstract}

\IEEEpeerreviewmaketitle

\section{Introduction}
\label{introduction}
\color{black}
The rapid advancements in modern technology, particularly within the domains of the \ac{IoT} and \ac{AI}, have catalyzed significant improvements in healthcare, enhancing both the quality and longevity of human life~\cite{qadri2020future,dimitrov2016medical,9294145}. An illustrative example of this transformative potential arises from the collaboration of researchers from Google Inc. and affiliated institutions,  wherein they showcased the remarkable potential of AI~\cite{gulshan2016development}. The study achieved diagnostic accuracy akin to physicians by training an AI system on an extensive dataset comprising tens of thousands of images. This proficiency extended to the precise identification of referable diabetic retinopathy while also revealing hitherto unrecognized correlations between distinct image patterns in fundus photographs and cardiovascular risk factors. Further innovation surfaces through ingeniously devised AI-powered \ac{IoT} systems, detailed in~\cite{sayeed2019neuro}, which meticulously track bodily movements, temperature fluctuations, and acoustic signals. This pioneering approach holds immense promise, particularly in promptly identifying the onset of epileptic seizures. Notably, these technological breakthroughs have been pivotal in addressing the challenges posed by the COVID-19 pandemic, where AI and IoT-driven solutions have played a crucial role in medical treatment and healthcare support~\cite{allam2020coronavirus}. Amidst this backdrop, medical data, including imaging and pathological data, are harnessed to train AI models capable of accurately identifying and predicting diseases, thereby augmenting the efficacy of healthcare diagnostics.

\indent The paradigm of data-driven medical applications has emerged as a promising avenue for creating robust and scalable predictive algorithms from medical data, garnering attention across academic and industrial circles ~\cite{cai2019survey}. Although these applications have undeniably improved healthcare quality~\cite{cascini2021developing}, their adoption is hindered by challenges related to security, privacy, and \ac{QoS}. In particular, concerns about data privacy loom large, with data owners, such as individuals, patients, and hospitals, apprehensive about sharing sensitive information with external entities. This apprehension has spurred the enactment of stringent legislation, exemplified by the United States Consumer Privacy Bill of Rights and the European Commission's \ac{GDPR}, aimed at safeguarding user privacy~\cite{andrew2021general,hartzog2020privacy,yuan2019policy,hoofnagle2019european}.

\begin{figure*}
	\centering
	\includegraphics[width= 6.5 in]{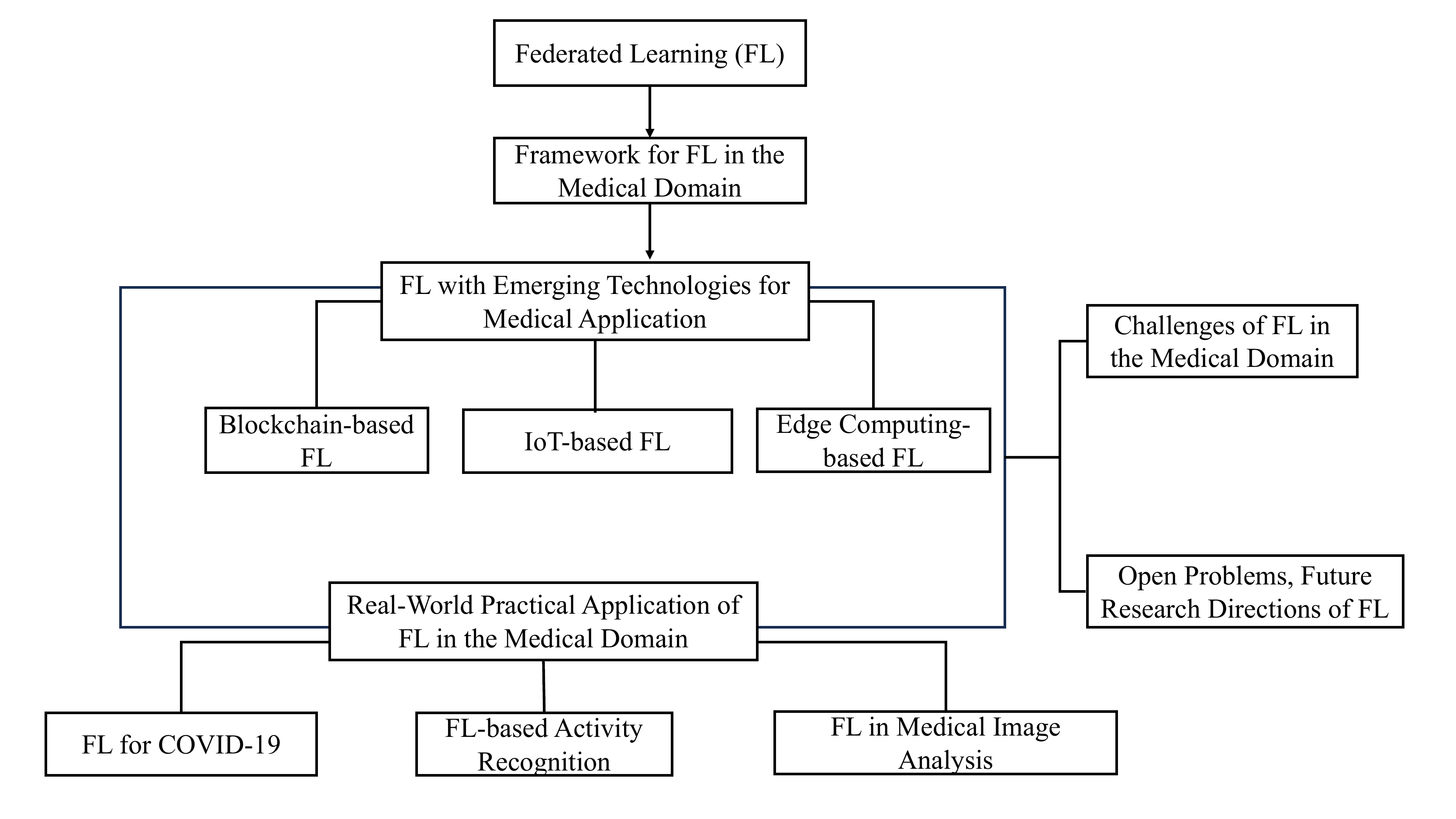}
	\caption{\color{black}Taxonomy of the topics covered in this work.}
	\label{taxonomy}
\end{figure*}

\begin{table}
\centering
\caption{Abbreviations used in our paper.}
\def\arraystretch{1.13}
\begin{tabular}{l|l}
\hline
\textbf{Acronym} & \hspace*{\fill} \textbf{Definition} \hspace*{\fill} \\ \hline
AI & Artificial Intelligence \\ \hline
ASD & Autism Spectrum Disorder \\ \hline
CFL & Clustered Federated Learning \\ \hline
CILL & Cyclic Institutional Incremental Learning \\ \hline
CNN & Convolutional Neural Networks \\ \hline
CPU & Central Processing Unit \\ \hline
CT & Computed Tomography \\ \hline 
DL & Deep Learning \\ \hline
DP & Differential Privacy \\ \hline
Edge-FL & Edge Computing-based Federated Learning \\ \hline
EHR & Electronic Health Record \\ \hline 
EMRs & Electronic Medical Records \\ \hline
FADL & Federated Autonomous Deep Learning \\ \hline 
FAIR & Findable, Accessible, Interoperable, Reusable \\ \hline
FATE & Federated AI Technology Enabler Framework \\ \hline
FATHOM & Federated Multi-task Hierarchical Attention Mode \\ \hline
FedAvg & Federated Averaging \\ \hline 
FL & Federated Learning \\ \hline
FML & Federated Machine Learning \\ \hline
FDL & Federated Deep Learning \\ \hline
FMs & Foundational Models \\ \hline 
GCAE & Generative Convolutional Autoencoder \\ \hline
GDPR & General Data Protection Regulation \\ \hline
GPU & Graphics Processing Unit \\ \hline
HIPAA & Health Insurance Portability and Accountability Act \\ \hline
I.I.D & Identical and Independently Distributed \\ \hline
IoMT & Internet of Medical Things \\ \hline
IoT & Internet of Things \\ \hline
IPFS & Interplanetary File System \\ \hline 
MEC & Mobile Edge Computing \\ \hline
ML & Machine Learning \\ \hline
MRI & Magnetic Resonance Imaging \\ \hline
NER & Named Entity Recognition \\ \hline
NLP & Natural Language Processing \\ \hline
Open-FL & Open Federated Learning \\ \hline
QoS & Quality of Service \\ \hline
TFF & Tensor Flow Federated \\ \hline
UCADI & Unified Computed Tomography AI Diagnostic Initiative \\ \hline
\end{tabular}
\end{table}
\indent 
The conventional approach of storing and processing data on distant cloud servers presents challenges in the context of medical data. Centralized \ac{ML} model training involves aggregating data onto a single machine or cluster, raising privacy concerns and practical limitations \cite{john2020developing,xu2020edge}. Furthermore, the trend is shifting towards distributed data storage and analysis at the network edge, driven by real-time requirements, latency concerns, and privacy considerations \cite{yu2017survey,chen2018thriftyedge,rafique2020complementing,sha2020survey,zhang2020stec,shi2016edge}.

In this landscape, the convergence of edge computing-based AI (a.k.a. EdgeAI) has emerged as a potent strategy~\cite{li2019edge,rausch2019towards,shi2020communication,ray2019edge}. EdgeAI leverages the processing capabilities of distributed devices, enabling localized \ac{AI} operations without violating data regulations. FL takes this a step further, allowing collaborative model training across multiple healthcare datasets without sharing sensitive patient data~\cite{rieke2020future,xu2021federated}. This approach addresses privacy concerns and boosts training efficiency by tapping into distributed datasets and resources. FL's potential is underscored by its applicability in healthcare, promising significant advancements in patient care and public health systems~\cite{luo2016big,lv2017next,maier2017surgical,rasouli2021artificial}.
\begin{table*}
\caption{Comparison and overview of recent surveys in FL.}
\label{tbl_related_works}
\centering
\begin{tabular}{ l|l}
 \midrule
 \textbf{Publications} & \hspace*{\fill} \textbf{Main Research Focus and Scope} \hspace*{\fill}  \\
 \midrule
 This Survey & \tabitem Presents a comprehensive survey on the use of FL in the medical domain. \\
 & \tabitem Provides a holistic taxonomy on the use of several technologies in conjugation with FL, especially for medical applications. \\
 & \tabitem Covers the state-of-the-art works targeting the most commonly caused cancer both in terms of incidence and mortality and the \\
 & \hspace{2.9ex} application of FL in developing computer-aided diagnosis tools. \\
 & \tabitem Discusses the main challenges, open research problems, and future research directions of FL in the context of the medical domain. \\ 
 \midrule
 Ref~\cite{abdulrahman2020survey} & \tabitem Focuses on comparing different ML-based deployment architectures for FL.   \\
 & \tabitem Covers privacy and security, and resource management.  \\
\midrule
 Ref~\cite{aledhari2020federated} & \tabitem Focuses on FL enabling technologies, protocols, and applications.  \\
 & \tabitem Provides a summary of the relevant FL protocols, platforms, and some real-life use-cases for FL.  \\
\midrule
 Ref~\cite{ji2021emerging} & \tabitem Focuses on exploring learning algorithms to improve the federated averaging algorithm. \\
 & \tabitem Reviews model fusion methods for FL. \\
 \midrule 
 Ref~\cite{li2020review} & \tabitem Summarizes the development prospects of FL in industrial field.\\
 \midrule
 Ref~\cite{li2021survey} & \tabitem Explores the definition of FL systems and analyzed the FL system components. \\
 & \tabitem Categorizes FL systems according to data distribution, ML model, privacy mechanism, communication architecture,  \\
 & \hspace{2.9ex}scale of federation and motivation of federation. \\
 \midrule
 Ref~\cite{lim2020federated} & \tabitem Surveys the applications of FL for mobile edge computing network optimization. \\
 \midrule
 Ref~\cite{01973} & \tabitem Investigates industrial application trends of FL, essential factors affecting the quality of FL models, and compares FL and \\
 & \hspace{2.9ex}non-FL algorithms in terms of learning quality. \\
 \midrule
 Ref~\cite{rieke2020future} & \tabitem Investigates FL as a potential solution for the future of digital health and highlights the challenges of FL on digital health.\\
 \midrule
 Ref~\cite{zheng2022applications} & \tabitem Discusses the application of FL for various fields in smart cities, including communication security and privacy issues. \\
 \midrule
 Ref~\cite{pfitzner2021federated} & \tabitem Provides a systematic literature review on the FL and its applicability for confidential healthcare datasets.\\
 \midrule
 Ref~\cite{xu2021federated,ali2022federated} & \tabitem Describes the role of FL in Internet of Medical Things (IoMT) networks for privacy preservation. \\
 \midrule 
 Ref~\cite{antunes2022federated} & \tabitem Presents a systematic literature review on current research about FL in the context of EHR data.\\
 & \tabitem Discusses a general architecture for FL based on the primary findings from the literature analysis.\\
 \midrule
 Ref~\cite{nguyen2022federated} & \tabitem Presents a survey on the use of FL in smart healthcare. \\
 & \tabitem Describes FL designs and some emerging applications of FL in healthcare domains.\\
 \midrule
\end{tabular}
\vspace{-5mm}
\end{table*}


\subsection{Scope and Contributions}
\textcolor{black}{The overarching goal of this survey is to explore the integration of FL with emerging technologies tailored for medical applications. While FL's versatility spans across multiple domains, its fusion with transformative innovations like IoT, blockchain, cloud and edge computing, and AI bears particularly impactful implications within the realm of healthcare. The safeguarding of health-related data poses distinct challenges owing to stringent regulations governing their acquisition and use. Even data anonymization falls short of guaranteeing privacy, as researchers in academia, industry, and health regulatory bodies concur that erasing patient metadata alone doesn't suffice. In light of these unique hurdles in medical data governance and privacy, our survey aspires to answer the following research questions (RQs):
\begin{itemize}
    \item \textbf{RQ1}. What constitutes the fundamental concept of FL, and what advantages does it offer within the realm of medical applications?
    \item \textbf{RQ2}. Within the healthcare domain, how do various FL frameworks differ, and what distinctive features characterize each of these frameworks?
    \item \textbf{RQ3}. What are the prevailing emerging technologies, and how is FL integrated with these emerging technologies to address healthcare challenges?
    \item \textbf{RQ4}. In practice, what are the tangible use cases of FL in the medical field?
    \item \textbf{RQ5}. Specifically, in the context of combating global cancer burdens, how is FL utilized to develop medical diagnostic tools? Is there empirical evidence supporting the superior performance and generalizability of FL compared to other data-driven AI models in healthcare?
    \item \textbf{RQ6}. Which open-source FL frameworks are currently available, and which among them is most suitable for deployment in medical settings?
    \item \textbf{RQ7}. What are the primary obstacles and significant challenges faced by FL when applied in healthcare contexts?
    \item \textbf{RQ8}. What open issues and future research directions emerge for FL within the medical domain?
\end{itemize}
Given the unique challenges in preserving privacy in medical data and based on the above-formulated RQs, our paper aims to provide a comprehensive taxonomy that clarifies the interplay between FL and emerging technologies in the healthcare sector. Furthermore, it explores the practical, real-world applications of FL in healthcare, delving into the challenges faced within this context and identifying open problems. Additionally, the paper outlines potential future research directions for FL in the medical field.}


Existing survey studies, exemplified by \cite{abdulrahman2020survey,aledhari2020federated,ji2021emerging,li2020review,li2021survey,lim2020federated,rieke2020future,zheng2022applications}, predominantly concentrate on facets such as general architecture, models, security, and privacy algorithms for FL that are unrelated to medical applications. Scarcely few, such as \cite{pfitzner2021federated,xu2021federated,ali2022federated,antunes2022federated,nguyen2022federated}, delve into FL within the medical domain, offering insights into works tailored for healthcare. However, the primary focus of our study rests on medical applications, where we underscore the weighty burden of
global cancer and illuminate the potency of FL in engendering
computer-aided diagnosis tools that address this challenge with
heightened efficacy. Thus, our survey's exclusive purview lies in scrutinizing FL's integration with emergent technologies dedicated to medical applications. Distinct from other FL-focused surveys, this article distinguishes itself by offering a comprehensive taxonomy encompassing the synergy of emerging technologies and FL, with a specific emphasis on medical applications. Fig.~\ref{taxonomy} illustrates the taxonomy of the subjects elucidated within this survey, while Table~\ref{tbl_related_works} provides a meticulous juxtaposition of related surveys in the field. In Section III, we expound upon FL's integration with emerging technologies within the healthcare context.


\indent The primary contributions of our work relative to the recent literature include:

\begin{itemize}
\item A comprehensive survey on the integration of FL with emerging technologies for medical applications, bridging the gap between technical rigour and healthcare context. 

\item Exploration of emerging technologies that synergize with FL to address medical challenges.

\item In-depth coverage of the recent works on the global cancer burden (targeting most commonly caused cancer in terms of incidence and mortality) and the application of \ac{FL} in developing computer-aided diagnosis tools. 


\item Overview and comparison of existing open-source FL software frameworks.

\item Identification of key challenges and open problems of FL within the medical domain.

\item Finally, this survey highlights the open problems of FL and offers future research directions to advance FL's role in healthcare.
\end{itemize}
\color{black}
\subsection{Paper Organizations}

The rest of the paper is organized as follows. Section~\ref{concepts_and_framework} briefly gives an overview of the FL concept and frameworks for FL in the medical domain. We explain the existing works in FL for medical applications in conjugation with other emerging technologies in Section~\ref{FL-Context-of-Healthcare}. {\color{black}In Section~\ref{pract}, we explain the practical application of FL in medical domain and illustrate how \ac{FL} is applied in developing medical diagnosis tools, particularly in addressing the global burden of cancer.} Open-source FL software frameworks are presented in Section~\ref{open_source_FL_Software}. We examine the challenges of FL from a medical perspective in Section~\ref{opportunities_and_challenges_FL}. Some open research problems and future research directions of FL in medical applications are presented in Section~\ref{Open_Problems_FL}. Finally, in Section~\ref{conclusion}, we provide concluding remarks.

\begin{figure*} [h!]
\centering
\includegraphics[width= 4.5 in]{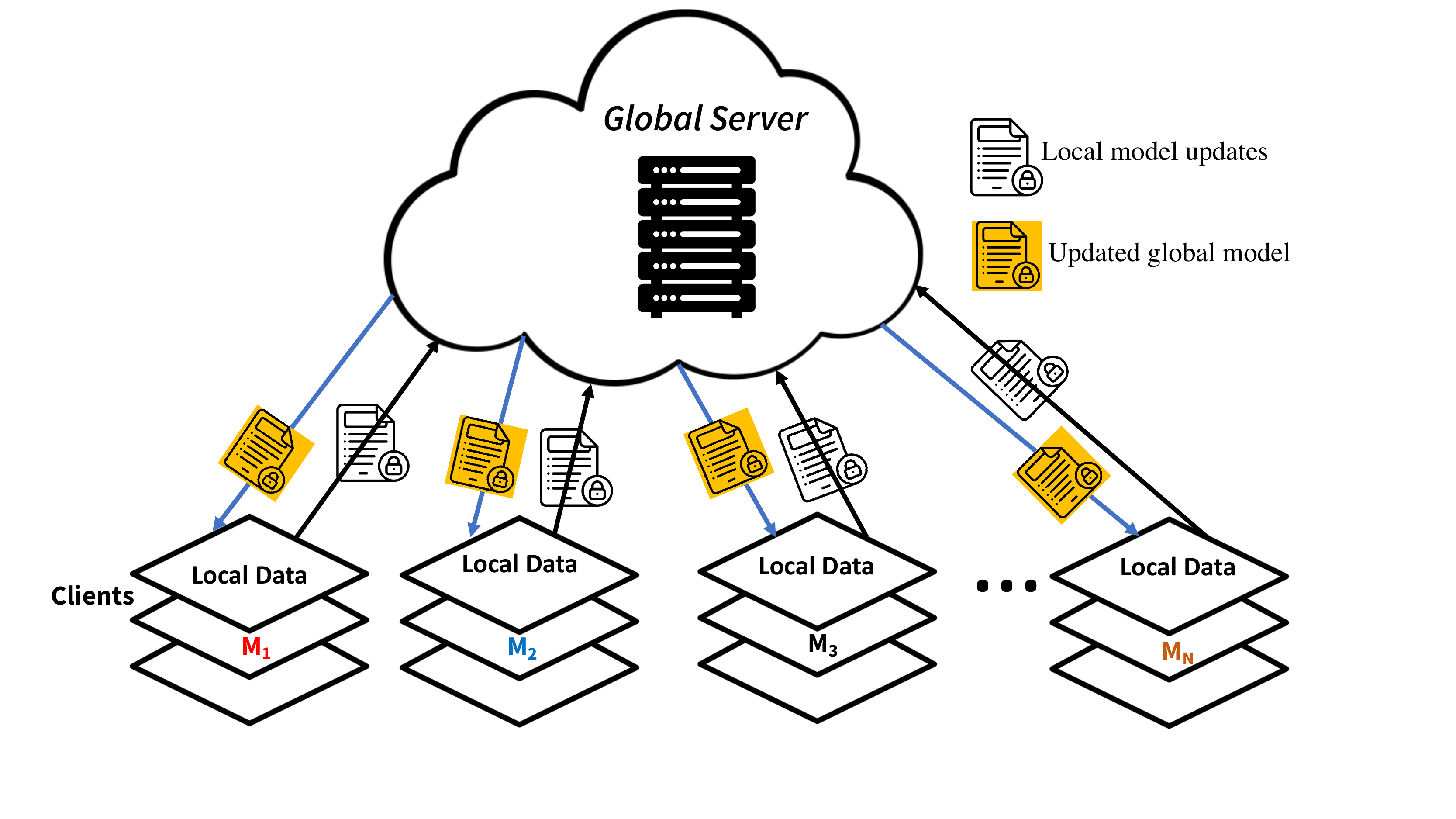}
\caption{Classical architecture of \ac{FL}. The clients send local model updates trained using the local dataset to the server for aggregation. The central server finally aggregates the local models transmitted by the participating clients and sends the current updated global model back to each participating client.}
\label {Federated_Learning_Architecture}
\end{figure*}

\begin{figure*} [htb]
\centering
\includegraphics[width= 4.5 in]{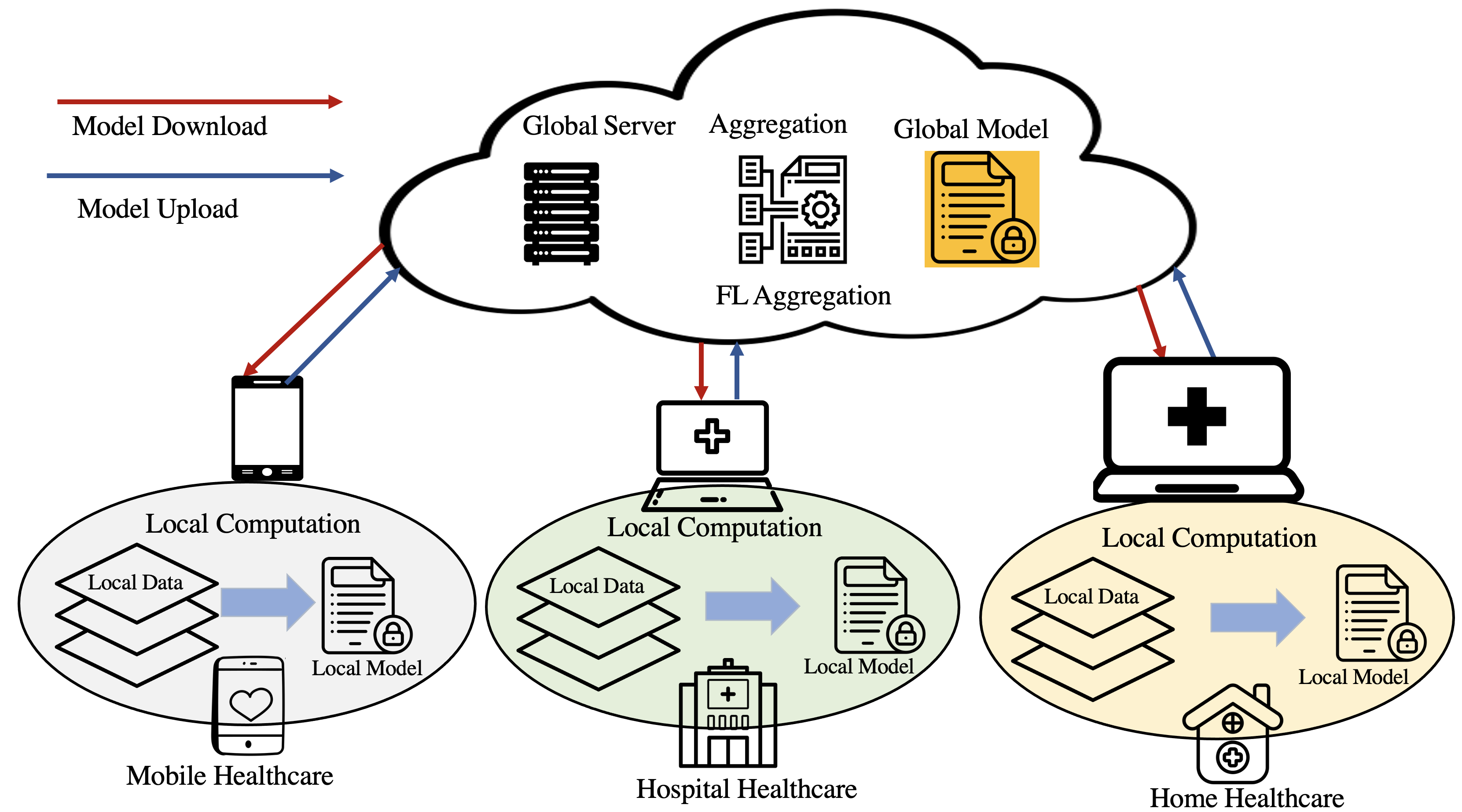}
\caption{FL framework for the medical domain.}
\label {FL_Framework}
\end{figure*}

\section{FL Concepts and Frameworks in the Medical Domain}
\label{concepts_and_framework}
\color{black}
Training complex ML models in a distributed medical environment is founded on the principles of trust, collaboration, efficiency, and scalability. In this paradigm, instead of centralizing all medical data on a single server for training, data remains on individual healthcare organizations or local hospital servers. This approach presents several key advantages:
\begin{itemize}
    \item \textbf{Privacy-Preserving}. Individual healthcare data remains decentralized, addressing concerns of data privacy and ownership. This is especially critical in medical applications where patient confidentiality is paramount.
    \item \textbf{Data Sovereignty}. Data remains under the control of its owners, which is crucial in sensitive domains like healthcare. Hospitals, clinics, and patients maintain authority over their respective data sources.
    \item \textbf{Reduced Data Transfer}. Since data doesn't need to be transmitted to a central server, bandwidth and latency issues are minimized, making it ideal for real-time applications such as immersive augmented reality and metaverse in healthcare.
    \item \textbf{Scalability}. Distributed training allows incorporating learning from a multitude of sources, enabling the creation of more robust and accurate models as the healthcare dataset size expands.
    \item \textbf{Energy Efficiency}. Healthcare devices in the network can contribute to model training while leveraging local processing power, reducing the need for massive data transfers.
\end{itemize}

Apart from the aforementioned advantages, in the healthcare domain, FL uniquely addresses privacy and security concerns by design:
\begin{itemize}
    \item \textbf{Data Localization}. FL operates on decentralized data sources, ensuring that sensitive medical information remains within the confines of its origin, reducing the risk of unauthorized access.
    \item \textbf{Data Aggregation}. Instead of sharing raw healthcare data, FL aggregates model updates or gradients from multiple devices, preserving individual healthcare data while generating collective insights.
    \item \textbf{Differential Privacy}. Techniques such as differential privacy can be integrated into FL, adding noise to aggregated updates to prevent the reconstruction of individual healthcare data.
    \item \textbf{Data Anonymization}. FL facilitates model training without requiring explicit data sharing, allowing health institutions to anonymize patient data and still contribute to model improvement.
    \item \textbf{Secure Communication}. Encrypted communication channels protect data during transmission between devices and the central server, mitigating interception risks.
\end{itemize}
In the context of medical applications, FL's role is pivotal. By enabling collaborative model training on decentralized data, FL optimally balances the need for improved healthcare insights with stringent privacy and security concerns. The approach ensures that sensitive patient information remains confidential while allowing the medical community to collectively advance diagnostic accuracy, treatment, and overall patient care.\\
\color{black}
\indent A classical \ac{FL} architecture is shown in Fig.~\ref{Federated_Learning_Architecture}. It consists of a centralized global server that broadcasts the AI/ML model parameters to the clients in the FL network. Clients are selected by the central server, either randomly or through a client selection algorithm~\cite{yu2020sustainable}. Upon receiving global model parameters, chosen clients proceed to train the model using their localized data. These clients then communicate their local model parameters back to the server, which aggregates and synthesizes them into a global model.\\
\indent A fundamental algorithm within the FL realm is \emph{FedAvg}, first introduced by Google~\cite{mcmahan2017communication}. The global server employs this algorithm for aggregating the local parameters of diverse clients in each iteration~\cite{xia2021auto,li2019convergence}. This iterative process continues until the desired convergence or a specified number of iterations is achieved on the global centralized server. While FedAvg effectively operates in the presence of non-iid data, its performance in non-iid data scenarios lacks theoretical guarantees within convex optimization settings~\cite{li2019convergence}.\\
\indent An analogous framework for FL within the medical domain is showcased in Fig.~\ref{FL_Framework}, wherein potential clients include mobile healthcare, hospital healthcare, and home healthcare. Notably, these domains may exhibit varying data distributions. The application of FL to these diverse domains hinges on their data characteristics to overcome security, privacy, healthcare system, and device challenges. For instance, mobile healthcare encompasses myriad smart healthcare devices, such as smartwatches, smartphones, and health monitoring gadgets, utilized by individuals for health tracking. Challenges encompass energy efficiency, communication, computation, and privacy considerations. Conversely, hospital healthcare involves larger institutions with substantial resources and computational capabilities, which mitigate concerns like client dropouts and straggler issues. Lastly, home healthcare involves medical services provided within residential settings, often comprising numerous smart healthcare devices. Remote patient monitoring falls under home healthcare, providing personalized care to patients in familiar surroundings. 

\color{black}
In Fig.~\ref{FL_Framework}, we emphasize the pivotal role of FL aggregation in orchestrating collaborative convergence from disparate data sources while upholding the sanctity of sensitive medical information. The aggregation process melds locally refined model updates or gradients from diverse decentralized devices or data sources. This integral step is central to fashioning an enhanced global model that elevates accuracy while steadfastly guarding data privacy and security.

Especially in the medical sector, where patient confidentiality is paramount, FL aggregation serves manifold purposes:
\begin{itemize}
    \item \textbf{Privacy Preservation}. FL aggregation guarantees the localization and security of individual patient data at its source, with the central server exclusively receiving aggregated model updates. This circumvents the need for raw data exchange, ensuring strict compliance with data protection regulations such as GDPR.
    \item \textbf{Collaborative Insights}. Aggregation brings together diverse datasets from various healthcare institutions or patient devices. This enables the generation of a more robust and accurate global model, which collectively learns from a broader spectrum of patient demographics and medical conditions.
    \item \textbf{Enhanced Diagnostics}. In the medical realm, the FL-aggregated model can yield improved diagnostic accuracy and predictive capabilities. For instance, a model trained on data from different hospitals can offer more generalized insights that benefit the entire healthcare community.
    \item \textbf{Customization}. FL aggregation allows for model customization while maintaining data security. Different institutions or regions can fine-tune the aggregated model based on local data nuances without directly sharing sensitive information.
    \item \textbf{Decentralized Expertise}. Aggregation facilitates the amalgamation of domain-specific expertise from various medical practitioners. This collective knowledge can lead to more accurate models that cater to specific medical conditions.
\end{itemize}

In essence, FL aggregation harmonizes the collective intelligence of individual data sources while preserving privacy, ensuring security, and yielding a globally refined model. This unified model, generated collaboratively from diverse medical data points, holds significant potential to drive advancements in medical research, diagnostics, and treatment while upholding the ethical imperatives of data privacy and security.
\color{black}

The categorization of FL frameworks within the medical domain is primarily determined by data distributions, as outlined below:

\subsection{Horizontal FL Framework}
Horizontal FL operates on datasets that share identical feature spaces across all devices. This implies that Medical Client A and Medical Client B possess the same set of features. For example, consider two regional healthcare institutions catering to separate user groups in distinct regions, yet having overlapping health diagnoses, resulting in identical feature spaces (e.g., age, gender, cholesterol levels, blood pressure)~\cite{lu2022personalized}. In this case, each hospital's data represents a horizontal partition of the overall dataset. The hospitals can collaborate using Horizontal FL to train a model that learns from the collective data without sharing individual patient records. The final model is a shared model that benefits from the diverse patient populations across different regions. Addressing limited sample sizes in data training, horizontal FL adopts a data-split approach. Li et al. exemplified this with an autism spectrum disorder (ASD) prediction scenario involving four medical institutions across diverse locations~\cite{li2020multi}. All these institutions share the same user features, and a global model is collaboratively trained within the FL framework using patient samples from all nodes.

\subsection{Vertical FL Framework}
Vertical FL entails collaborative training of a global model from varied feature spaces using diverse datasets~\cite{romanini2021pyvertical}. This approach increases the feature dimension during data training. Consider a large hospital and a medical insurance company coexisting in a location; they might share a common user database. However, their feature spaces could significantly differ, with the hospital focusing on patient treatment history and the insurance institution managing medical bills, receipts, and claims. Vertical FL capitalizes on these distinct datasets to offer location-specific health recommendations. Vertical FL allows these medical institutions to merge their data for a holistic analysis without compromising data privacy. They exchange only the relevant features (e.g., combining demographic data with genetic markers) needed for a particular research task, such as identifying genetic predispositions to certain medical conditions. This approach ensures that sensitive patient information remains confidential. An example is provided by Cha et al., where an autoencoder FL model transforms client user features into a latent dimension for vertically partitioned medical data~\cite{cha2021implementing}.

\subsection{Federated Transfer Learning Framework}
Federated transfer learning emerges when vertical FL incorporates a pre-trained model, initially trained on a similar dataset to address a distinct problem. This strategy is especially useful when datasets differ not just in samples but also in feature spaces. It facilitates training a customized health model tailored to user-specific attributes. Specifically, in this framework, medical institutions collaborate by not only sharing model updates but also transferring knowledge gained from related tasks, enhancing the performance of their models. Consider a consortium of healthcare institutions aiming to build a personalized cancer prediction model. Each institution has access to data on different cancer types (e.g., breast cancer, lung cancer) and wants to leverage the insights gained from one type to improve predictions for another while maintaining data privacy.  In Federated Transfer Learning, institutions initially train models independently for their specific cancer types. Afterwards, they share model updates and insights with the central model, allowing it to benefit from the collective knowledge. For instance, discoveries made during breast cancer prediction can be transferred to improve the lung cancer prediction model and vice versa. This approach optimizes model performance and generalization without compromising data confidentiality. An illustrative instance is \emph{FedHealth} proposed by Chen et al., which aggregates data from diverse organizations~\cite{chen2020fedhealth}. \emph{FedHealth} creates personalized models for each organization using federated transfer learning. The FL model initially learns human activity recognition tasks and then employs transfer learning to extend this classification to categorize Parkinson's disease. This approach generates a global model for a specific disease prediction that can also be applied to other medical challenges.




\color{black}
\section{FL and Its Applications with Other Emerging Technologies in the Context of Healthcare}
\label{FL-Context-of-Healthcare}

FL stands out with its unique ability to forge robust and dependable ML models, all without necessitating the sharing of raw data. Its potential applications span diverse domains such as healthcare~\cite{rieke2020future, brisimi2018federated, huang2019patient}, \ac{NLP}~\cite{lin2021fednlp}, transportation\cite{liu2020privacy, tan2020federated}, and finance~\cite{cheng2020federated}. In an era characterized by the proliferation of networked \ac{IoT} devices, amassing and securing the deluge of data they generate pose substantial challenges. The conundrum of preserving the confidentiality and privacy of medical records, in particular, has become a focal point for researchers spanning industry, academia, and medical research. Within this landscape, FL techniques have emerged as indispensable privacy preservation tools in both industrial and medical AI applications. Recent studies underscore the efficiency gains and relief they offer to global healthcare systems~\cite{shen2019medchain}. Furthermore, FL models exhibit heightened robustness and efficacy compared to traditional data-driven medical applications~\cite{grama2020robust, fang2020local}.

To broaden its horizons and expand into diverse application domains, FL seamlessly collaborates with other cutting-edge technologies. This synergy accelerates the collective training of ML and DL models without necessitating the centralization of data sharing. In the ensuing sections, we delve into how FL harmoniously interfaces with contemporary technologies, with a specific focus on its application within the medical domain.

\subsection{Blockchain-based FL in Healthcare}
\label{blockchain-based-FL-in-Healthcare}

FL is heralded for safeguarding the privacy of raw data on individual clients while enabling collaborative model development, as depicted in Fig.~\ref{Federated_Learning_Architecture}. However, inherent challenges persist (refer to Section~\ref{opportunities_and_challenges_FL}). To surmount these hurdles and augment FL's security, scalability, and performance, a powerful ally emerges in the form of blockchain technology~\cite{bao2019flchain, desai2021blockfla, ramanan2020baffle}.

At its core, blockchain is a decentralized, public ledger technology fostering collaborative learning across devices without reliance on a central aggregator~\cite{bao2019flchain, desai2021blockfla, ramanan2020baffle}. Its versatility spans a gamut of data-centric domains, with healthcare being a prominent arena. While originally devised for managing Bitcoin transactions and cryptocurrencies~\cite{nakamoto2008bitcoin}, blockchain's potential stretches far beyond. It has been instrumental in reshaping various sectors, including business, transportation, logistics, and healthcare~\cite{frizzo2020blockchain, pournader2020blockchain}. In healthcare, blockchain's capabilities offer secure management of electronic health records~\cite{engelhardt2017hitching} and have prompted extensive research exploration~\cite{agbo2019blockchain, griggs2018healthcare, holbl2018systematic, qadri2020future, zubaydi2019review, hasselgren2020blockchain, zubaydi2019review, vazirani2020blockchain}. The amalgamation of blockchain and FL provides an avenue for secure health data storage and efficient deployment of FL applications in the healthcare sector.

In this symbiotic alliance, blockchain enhances FL's robustness and privacy. Pioneering endeavors like the one outlined in~\cite{passerat2019blockchain} showcase a secure, decentralized architecture employing privacy-preserving encryption techniques for FL within healthcare, leveraging the Ethereum blockchain. This architecture logs network events while preserving patient identities through advanced encryption protocols. Such secured frameworks hold immense potential for privacy-conscious AI applications in healthcare~\cite{passerat2019blockchain}.

The amalgamation of decentralized data and collaborative FL approaches~\cite{bonawitz2017practical} aligns seamlessly with the decentralized nature of blockchain and can expedite the development of AI applications in healthcare. For instance, the architecture detailed in~\cite{polap2021agent} combines blockchain and FL to secure multi-agent systems, promoting cooperation among individual agent units for \ac{IoMT}. The results demonstrate significant promise, achieving an 80\% accuracy in skin cancer classification~\cite{polap2021agent}.

Blockchain's potential extends to clinical trials and precision medicine~\cite{shae2018transform}, offering a distributed parallel computing architecture for precision medicine through FL and transfer learning~\cite{shae2018transform}. This innovation facilitates decentralized processing of extensive medical data. Disease diagnosis also benefits from blockchain; Health-Chain~\cite{chen2019asynchronous}, for instance, presents a blockchain-based decentralized privacy-preserving cross-institution disease classification framework. Employing differential privacy and pseudo-identity mechanisms, it addresses data privacy issues effectively. The authors' experiments, focusing on breast cancer diagnosis and ECG arrhythmia classification, demonstrate the efficiency and effectiveness of Health-Chain~\cite{chen2019asynchronous}.

In essence, the symbiotic integration of blockchain-based FL in healthcare holds significant promise for the industry. Here are some key points highlighting its importance:
\begin{itemize}
    \item \textbf{Enhanced Data Security}. Blockchain's inherent security features, such as decentralized and immutable ledgers, bolster the privacy and integrity of healthcare data. When combined with FL, this integration ensures that patient information remains highly secure.
    \item \textbf{Data Transparency and Accountability}. Blockchain's transparency allows patients to have greater control and visibility over who accesses their health data and for what purposes. FL, in tandem, maintains data privacy while providing an auditable and transparent record of model updates and access requests, enhancing trust among stakeholders.
    \item \textbf{Interoperability}. Blockchain facilitates interoperability among disparate healthcare systems and institutions. FL's federated approach extends this interoperability to machine learning models, enabling seamless collaboration and knowledge sharing across healthcare providers and researchers.
    \item \textbf{Consent Management}. Blockchain-based smart contracts can manage patient consent for data sharing and model training. FL can then ensure that only authorized parties access and utilize patient data, aligning with evolving data protection regulations.
    \item \textbf{Data Monetization}. Patients can potentially benefit from sharing their health data through blockchain-based tokens or incentives. FL ensures data privacy while allowing patients to have control over how their data is used, possibly leading to new revenue-sharing models.
    \item \textbf{Research Advancements}. The combination of blockchain and FL accelerates medical research by facilitating secure and collaborative model training across institutions and geographies, leading to the development of more accurate diagnostic and treatment tools.
    \item \textbf{Regulatory Compliance}. Healthcare is heavily regulated, and blockchain-based FL can assist in complying with data protection laws like HIPAA or GDPR by providing a robust framework for data management and privacy.
\end{itemize}
In summary, the integration of blockchain and FL in healthcare not only addresses critical data privacy concerns but also enhances data sharing, research capabilities, and patient empowerment, ultimately driving innovation and improving patient care.

\subsection{FL Enabled IoT-based Healthcare Monitoring}
\label{FL-Cloud-Computing-IoT-Healthcare-Monitoring}

The advent of \ac{IoT} technology has orchestrated a revolutionary landscape, intertwining seamless connectivity, smart devices, and enhanced productivity~\cite{care2016internet}. This evolution has been steered by the integration of a multitude of distributed smart devices and sensors, orchestrating the real-time generation of user data across various applications. Among the diverse domains that have been profoundly impacted by the \ac{IoT} revolution, healthcare stands out prominently~\cite{dang2019survey}. Combining IoT with FL in healthcare merges real-time data from IoT devices with privacy-preserving FL model training. Notably, the work outlined in~\cite{dang2019survey} delves into the intricacies of this fusion, elucidating the interplay between \ac{IoT} and healthcare applications. This exploration is complemented by an insightful discussion on the security challenges encountered within this convergence and potential strategies to address them.

The \ac{IoMT} paradigm represents yet another transformative facet within the realm of IoT and healthcare technology~\cite{wang2022privacy}. As these innovative technologies continue to proliferate, they give rise to a profusion of data streams from integrated devices. However, a distinctive feature emerges-thanks to \ac{FL} techniques-allowing data evaluation to transpire locally at the edge devices. This decentralized approach, a cornerstone of \ac{FL}, ensures data privacy for \ac{IoT} devices, particularly critical within sensitive domains such as healthcare. It distinguishes \ac{FL} from conventional centralized \ac{ML} methods, promising a balance between data utility and confidentiality. In a parallel vein, Zhao et al.~\cite{zhao2020privacy} presents a pioneering endeavor that harnesses \ac{FL}, differential privacy, and blockchain to empower manufacturers to assess \ac{IoT}-derived data securely and effectively, aligning with the narrative of enhanced healthcare monitoring.

The confluence of \ac{IoT} and \ac{FL} holds great potential, fundamentally transforming healthcare and beyond. Here are some key points highlighting its importance:

\begin{itemize}
    \item \textbf{Real-time Data Processing}. IoT devices continuously collect and transmit patient health data, enabling healthcare providers to access real-time information. FL can be used to collaboratively analyze this data while preserving patient privacy, leading to timely insights and interventions.
    \item \textbf{Remote Patient Monitoring}. IoT devices can remotely monitor patients' vital signs and chronic conditions. FL allows this data to be aggregated and analyzed collectively, enabling healthcare professionals to make informed decisions about patient care without compromising individual privacy.
    \item \textbf{ Early Disease Diagnosis and Prognosis}. By analyzing IoT-generated data with FL models, healthcare providers can predict and detect diseases at an early stage. This can lead to timely interventions and improved patient outcomes.
    \item \textbf{Reduced Healthcare Costs}. The integration of IoT and FL can help reduce healthcare costs by preventing hospital readmissions through continuous monitoring, optimizing treatment plans, and reducing unnecessary medical interventions.
    \item \textbf{Scalability}. The IoT ecosystem is continually expanding, and FL can scale to accommodate a growing number of IoT devices and data sources, making it suitable for large-scale healthcare applications.
    \item \textbf{Improved Patient Engagement}. IoT devices can engage patients in their healthcare by providing them with valuable insights into their health status. FL can further enhance this engagement by offering personalized recommendations and feedback.
\end{itemize}

In summary, IoT devices collect patient data, while FL enables collaborative model building across decentralized sources without sharing raw data. This empowers personalized treatment recommendations, real-time monitoring, and alerts. Patient privacy is upheld as data remains local, meeting stringent regulatory standards. Moreover, diverse IoT data enriches models, adapting them to changing conditions and ensuring scalability. This convergence, as evidenced by Zhao et al.~\cite{zhao2020privacy}, holds the promise of revolutionizing data evaluation methodologies, ensuring privacy, security, and progress in the context of medical applications.

\subsection{Edge Computing Assisted FL for Healthcare} 
\label{edge_computing_FL}

\begin{figure} [!t]
\centering
\includegraphics[width=\columnwidth,scale=0.99,keepaspectratio]{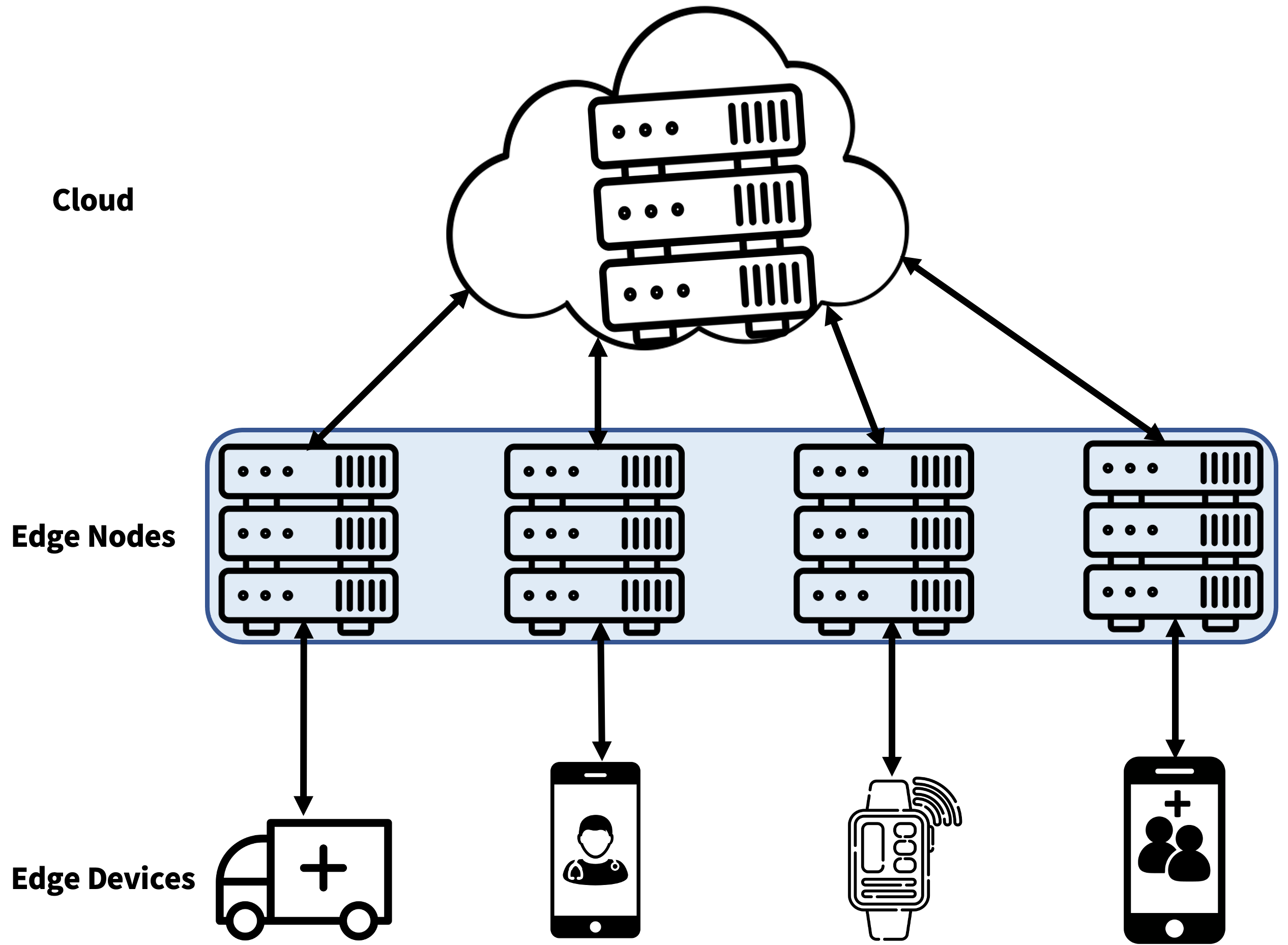}
\caption{General architecture of edge computing in healthcare.}
\label {edge_computing}
\end{figure}

In the ever-evolving landscape of healthcare, the fusion of edge computing and FL emerges as a transformative force. Edge computing extends the realms of conventional cloud computing by relocating computing resources and data storage closer to data sources, ushering in improvements in network resilience, availability, latency, and bandwidth~\cite{abreha2021monitoring, mao2017survey, mach2017mobile}. Fig.~\ref{edge_computing} elucidates the architecture of an edge computing platform, decentralizing processing power, intelligence, and communication capabilities to the myriad edge devices. This shift substantially reduces response times and curtails the deluge of data streaming onto the internet, a boon for time-sensitive applications.

A pioneering FL framework, FedHome~\cite{wu2020fedhome}, exemplifies the integration of cloud and edge resources to personalize in-home health monitoring. FedHome refines models by employing a class-balanced dataset generated from user data, effectively addressing statistical and communication challenges. It employs a lightweight model, the \ac{GCAE}, for seamless data transfer between the cloud and the edges, significantly reducing communication costs. This remarkable integration yields an impressive accuracy of 95.41\%, outperforming conventional neural network techniques by over 7.49\%. Compared to other FL systems, FedHome achieves accuracy improvements exceeding 10\%, positioning it as a potent tool for future in-home health monitoring~\cite{wu2020fedhome}.

In parallel, innovative approaches continue to surface. A fog computing-based prototype~\cite{garcia2020accelerating} demonstrates a fourfold acceleration in the computation process for mobile patients. Another trailblazing framework, BodyEdge\cite{pace2018edge}, comprises a miniature mobile client module on mobile devices and an edge gateway, featuring a multi-radio interface for dependable connectivity and multi-technology communication. Designed to thrive in local environments, this healthcare framework significantly reduces data traffic to the internet, achieving response times commensurate with healthcare IoT applications\cite{pace2018edge}.

FL, with its innate compatibility with edge computing, holds substantial promise. Unlike traditional centralized approaches, edge computing obviates the need to transport data from distributed edge devices to a central cloud platform. This synergy leverages the computational resources of edge servers and the data harnessed from edge devices, a match made in healthcare innovation heaven. Personalized edge-assisted FL~\cite{hakak2020framework} offers a prime example, catering to healthcare analytics rooted in user-generated data. This framework employs pre-trained models to glean tailored insights, examining diseases by monitoring mobility levels and behaviors through wearable devices. The result is a privacy-respecting and cloud-resource-efficient solution\cite{hakak2020framework}.

To bolster integrity and security in the realm of distributed IoT devices, the fusion of FL and blockchain within edge computing architectures comes to the fore~\cite{nguyen2021federated}. Recent advancements in blockchain and \ac{MEC} herald fresh opportunities for healthcare transformation~\cite{rahman2018blockchain, asif2018toward, abdellatif2020sshealth, rahman2018spatial, kuo2017blockchain}. Decentralized architectures, like the one proposed in~\cite{nguyen2020blockchain}, employ blockchain and MEC for secure \ac{EMR} sharing among federated hospitals. Such systems deploy decentralized EMR storage on MEC servers, supporting secure, distributed data sharing through the \ac{IPFS}. For energy-efficient computation offloading decisions within MEC systems, deep reinforcement learning techniques and FL frameworks with mobile edge systems step in, conserving IoT device energy while ensuring service quality~\cite{wang2019edge}.

Ensuring the sanctity of personal data in smart healthcare systems is paramount. The lightweight privacy protection protocol presented in~\cite{wang2022privacy} operates within an edge computing environment, leveraging shared secrets and weight masks to safeguard user privacy. This protocol maintains gradient privacy, fortifying model accuracy while repelling equipment dropouts and collusion attacks. It employs algorithms rooted in digital signatures and hash functions, preserving message integrity and thwarting replay attacks. When compared to differential privacy, this approach boasts a 40\% efficiency improvement while upholding the same safety and efficacy standards as FL~\cite{wang2022privacy}.

At this juncture, we want to highlight that the symbiotic integration of Edge Computing-based FL in healthcare is a transformative development with several key points of significance:
\begin{itemize}
    \item \textbf{Low Latency Data Processing}. Edge computing brings computation and data processing closer to the source of data, reducing latency. When combined with FL, this enables real-time analysis of healthcare data from IoT devices, wearables, and sensors, ensuring timely responses to critical medical conditions.
    \item \textbf{Privacy-Preserving Data Processing}. Edge devices can locally process and aggregate data, allowing FL to perform model training without the need to centralize sensitive patient information. This preserves patient privacy and complies with data protection regulations.
    \item \textbf{Enhanced Data Security}. Edge devices can implement security measures locally, protecting data as it is collected and processed. FL extends this security to model training, ensuring that patient data remains secure throughout the entire process.
    \item \textbf{Edge AI for Diagnosis}. Combining Edge Computing with FL enables AI-powered diagnosis and decision support at the edge. This can lead to faster and more accurate diagnosis in scenarios like medical imaging analysis.
    \item \textbf{Edge Devices Diversity}. Healthcare relies on a diverse range of devices, from wearable fitness trackers to sophisticated medical equipment. Edge-based FL can adapt to this diversity, providing a flexible platform for healthcare applications.
    \item \textbf{Cost-Efficiency}. Edge computing can reduce the costs associated with data transfer and centralized cloud infrastructure. FL's decentralized training further optimizes resource usage.
\end{itemize}
In summation, the integration of edge computing and FL unlocks new horizons in healthcare, driving innovation by reducing latency, conserving bandwidth, and safeguarding data privacy.  This powerful synergy has the potential to revolutionize healthcare delivery, especially in remote and resource-constrained settings.

\section{Practical Application of FL in Medical Domain}
\label{pract}
In this era of data-driven decision-making, the convergence of AI and healthcare holds tremendous promise. Among the various AI methodologies, FL emerges as a trailblazing approach that marries the potential of machine learning with the sensitivity of medical data. For example, medical institutions across the globe collaborate seamlessly, sharing insights and expertise, without compromising the privacy and security of patient records. FL makes this vision a reality by revolutionizing how healthcare organizations leverage data. It ushers in an era where medical breakthroughs are a collective endeavor and where the protection of individual privacy is of utmost importance. In this section, we embark through the practical applications of FL in the medical domain, exploring how this groundbreaking technology is reshaping patient care, research, and healthcare systems worldwide. 

\color{black}
\subsection{FL-enabled COVID-19 Detection}
The realm of healthcare witnessed a transformative shift with the advent of smart healthcare and IoMT technologies. Amid global health crises such as the COVID-19 pandemic, the need for rapid, collaborative responses became paramount. FL, known for its privacy-preserving capabilities, emerged as a powerful tool in the fight against the pandemic. Researchers have harnessed FL's potential to contribute to the analysis of the COVID-19 outbreak, leading to a wealth of innovative applications~\cite{dayan2021federated, xu2020collaborative, raisaro2020scor, vaid2021federated,yang2021federated, dou2021federated}. A comprehensive review of FL in COVID-19 detection can be found in~\cite{naz2022comprehensive}, while~\cite{qian2021value} offers insights into recent FL applications, limitations, and challenges in both COVID-19 and non-COVID-19 scenarios. Specifically, how the FL could provide valuable care during and post-COVID-19 was a question of interest.

\indent Realizing the lack of generalization in AI models, which prevents them from being used in clinical settings, one of the pioneering applications of FL in healthcare is the Unified \ac{CT}-COVID AI Diagnostic Initiative (UCADI)~\cite{xu2020collaborative, bai2021advancing}. UCADI facilitates global collaboration in building a clinic CT-COVID AI application. By uniting efforts worldwide, it leverages diverse data sources for enhanced diagnostic accuracy while preserving privacy. FL has also shown promise in predicting mortality among hospitalized COVID-19 patients~\cite{vaid2021federated, vaid2020federated}. By aggregating patient data from multiple hospitals within a health system, FL models outperformed locally trained models in predicting outcomes, demonstrating the potential to develop effective predictive tools without compromising patient privacy. Another groundbreaking initiative, the \ac{EMR} Chest X-ray AI model (EXAM)~\cite{dayan2021federated}, utilizes FL to estimate future oxygen requirements for symptomatic COVID-19 patients. Drawing data from 20 global institutes, EXAM leverages vital signs, laboratory data, and chest X-rays to improve patient care. Collaborative FL frameworks enable medical institutions to screen COVID-19 from chest X-ray images without sharing patient data~\cite{feki2021federated}.  Numerous studies have proposed and verified FL's effectiveness in COVID-19 X-ray data training and deployment experiments~\cite{yan2021experiments, liu2020experiments, zhang2021dynamic, abdul2021covid, bhattacharya2022application, ho2021dpcovid, de2022fully, cetinkaya2021communication, ulhaq2020covid, banerjee2020multi, alam2021federated, slazyk2022cxr, cao2021near, boyi2020experiments}.

\indent Hybrid framework for COVID-19 prediction via federated \ac{ML} models is investigated in~\cite{kallel2022hybrid}. An asynchronously updating FL model for mobile and deployable resource nodes to create local AI models for COVID-19 detection without explicit radiograph data exchange with the cloud is proposed in~\cite{sakib2021covid}. In~\cite{qayyum2021collaborative}, Qayyum et al. utilized the emerging concept of clustered FL (CFL) for an automatic diagnosis of COVID-19  at the edge by training a multi-modal ML model capable of detecting COVID-19 in both X-ray and Ultrasound imagery. The authors also envisaged that CFL is found to cope better with the divergence in data distribution from different sources than the conventional FL technique. Similarly, in \cite{chen2021multi}, Chen et al. discussed the multi-modal COVID-19 discovery based on collaborative FL. Moreover, in~\cite{jaladanki2021development}, FL was used to predict acute kidney injury within three and seven days of admission in 4029 adults hospitalized with COVID-19 at five socio-demographically diverse New York City hospitals using demographics, comorbidities, vital signs, and laboratory values. The prediction performance of FL models was often found to be higher than that of single-hospital models and comparable to that of pooled-data models. 

\indent A 5G-enabled architecture of auxiliary COVID-19 diagnosis based on FL for various institutions and central cloud cooperation was proposed by Wang et al. in order to realize the sharing of diagnosis models with high generalization performance~\cite{wang2021auxiliary}. A COVID-19 diagnosis model cognition framework was built on sharing and updating the model adaptively between the central and distributed nodes to interchange models and parameters. Federated \ac{DL} for detecting COVID-19 lung abnormalities in CT through a multinational validation study was conducted in \cite{dou2021federated}. The authors investigated FL strategies for developing an AI model for COVID-19 medical image diagnostics with good generalization capabilities on unseen multinational datasets. A secure international informatics FL infrastructure to investigate COVID-19 is initiated in \cite{raisaro2020scor}. To address the privacy dilemma for COVID-19 data, the SCOR consortium has been formed that has created a ready-to-deploy secure FL infrastructure based on privacy and security technology. For COVID-19 impacted region segmentation in 3D chest CT, the authors in \cite{yang2021federated} presented a federated semi-supervised learning architecture using multinational COVID-19 data from China, Italy, \& Japan. The proposed approach can extract useful information from clients who only have unlabeled data.

\color{black}
\indent These applications exemplify how FL has emerged as a critical tool in the medical domain, offering global collaboration, predictive power, and privacy preservation in the fight against COVID-19 and beyond. During the pandemic, through FL, medical institutions worldwide could share insights and collectively train predictive models for virus spread, vaccine development, and treatment strategies while safeguarding patient information. FL's decentralized nature allowed for real-time data analysis and rapid response. Healthcare professionals and researchers could continuously update and refine models as new data became available, leading to more accurate predictions and recommendations. Additionally, FL facilitated global collaboration, enabling experts from various regions to pool their data and expertise to combat the virus effectively. This collaborative approach was instrumental in accelerating vaccine research and development, as researchers worldwide could collectively analyze clinical trial data without compromising data privacy. Overall, FL emerged as a crucial tool in the fight against COVID-19 by enabling privacy-preserving data collaboration, real-time analysis, and global cooperation among healthcare professionals and researchers.

\subsection{FL-enabled Activity Recognition for Elderly Care}

The realm of human activities encompasses a diverse spectrum, ranging from sedentary behaviors such as sitting and standing to dynamic actions like walking and running. The domain of activity recognition, driven by computer vision, holds immense potential to revolutionize elderly care. It not only facilitates preventive and predictive interventions but also paves the way for personalized care through intelligent fall detection solutions, as demonstrated in the work by Brenvcivc et al.~\cite{brenvcivc2020intuitive}. Within this context, \ac{FL} emerges as a compelling arena for advancing activity recognition's role in elderly care, ushering in solutions tailored for aging populations while meticulously safeguarding their privacy.

A pivotal study conducted by Sozinov et al.~\cite{sozinov2018human} illuminates the landscape of FL's contribution to activity recognition. Their work compared the performance of FL against centralized learning methodologies, employing deep neural network architectures and softmax regression models. The investigation spanned both synthetic and real-world datasets, delving into crucial aspects such as transmission costs and the influence of erroneous clients with compromised data. The findings unveiled a nuanced scenario—while FL models exhibited slightly diminished accuracy compared to their centralized counterparts (achieving up to 89\% accuracy as opposed to 93\% in centralized models), the outcomes remained notably acceptable. This study underscores the potential of FL to deliver proficient models for human activity recognition tasks. Stepping further, the pursuit of efficient feature representation finds its expression in the innovative Federated Multi-Task Hierarchical Attention Model (FATHOM) proposed by Chen et al.~\cite{chen2019federated}. This novel framework intertwines classification and regression models from diverse sensors through a federated approach. By addressing the intricacies of multi-sensor data, FATHOM aspires to enhance the accuracy of activity recognition while embracing the distributed nature of FL. The architecture introduced by Bram et al.~\cite{bramfederated} introduces a powerful \ac{FML} software framework designed to harness the potential of deep neural networks. This architecture gracefully accommodates vast quantities of unlabeled data often dispersed across distributed clients, leveraging FL's decentralized training paradigm to yield various neural network architectures. Intriguingly, FL's applicability extends to the realm of elderly populations grappling with Alzheimer's and dementia, as explored in Hesseberg's work~\cite{hesseberg2020federated}. This study delves into FL's potential to cater to the unique challenges faced by this demographic, opening avenues for enhancing care and understanding within the context of cognitive impairments.

In the realm of activity recognition for enhanced elderly care, FL emerges as a potent tool, ushering in solutions that resonate with privacy-centric, personalized care paradigms. The studies elucidated here underscore FL's capacity to augment accuracy, feature representation, and interdisciplinary collaborations, rendering it a pivotal catalyst in redefining elderly care in the era of advanced technology.

{

\color{black}
\subsection{Practical Use Cases of FL for Medical Diagnostic Tools}
\label{practical_use_cases}

Here, we will illustrate how \ac{FL} is applied in developing medical diagnosis tools, particularly in addressing the global burden of cancer and other non-communicable diseases.

\ac{FL} is being increasingly explored in medical image analysis to train the \ac{DL} models on the large datasets distributed across multi-center and cross-border~\cite{roth2020federated, sheller2020federated, remedios2020federated}. It allows each medical center to train a global model collaboratively in FL settings. Cross-domain collaborative and decentralized learning has several advantages. In the field of medical imaging, data annotation is one of the crucial and labour-intensive tasks. Using \ac{FL}, different institutions can benefit from each other's annotations without even sharing them. Training \ac{DL} algorithms require high computational power and memory space. Using \ac{FL} can help in efficient training and memory consumption for AI-assisted medical image analysis algorithms. In this Section, we will briefly outline the existing works on different types of medical use cases and their application using \ac{FL}. In this regard, we will cover examples of the most commonly occurring cancer in terms of incidence and the most deadly cancer in terms of mortality. We also discuss the application of \ac{FL} in developing the computer-aided diagnosis tool for those medical cases. The most frequently occurring cancer in terms of incidence are breast, lung, colorectum, prostate, stomach, liver, etc. (please refer to Fig.~\ref{fig:cancerstatistics1}). The most lethal cancers are lung cancer, colorectum cancer, liver cancer, stomach cancer, breast cancer, esophagus cancer, pancreas, etc.~\cite{sung2021global} (see Fig.~\ref{fig:cancerstatistics2}). Below, we briefly describe the most commonly caused cancer both in terms of incidence and mortality.  

\vspace{1.0ex}
\noindent \textbf{Glioblastoma tumor detection}. Researchers from a medical center at the University of Pennsylvania, Penn Medicine, and Intel Labs have used \ac{FL} to develop a model that can detect glioblastoma tumors in \ac{MRI} images with 33\% higher accuracy than traditional methods~\cite{pati2022federated}. The model was trained on data from 71 sites across 6 continents, without any of the data ever leaving the individual institutions. This allowed the researchers to build a more robust and generalizable \ac{FL} model that can be used to diagnose glioblastoma in patients around the world.

\begin{figure} [!t]
    \centering
    \includegraphics[trim=2cm 0.5cm 2.5cm 1cm, clip,width=0.9\linewidth]{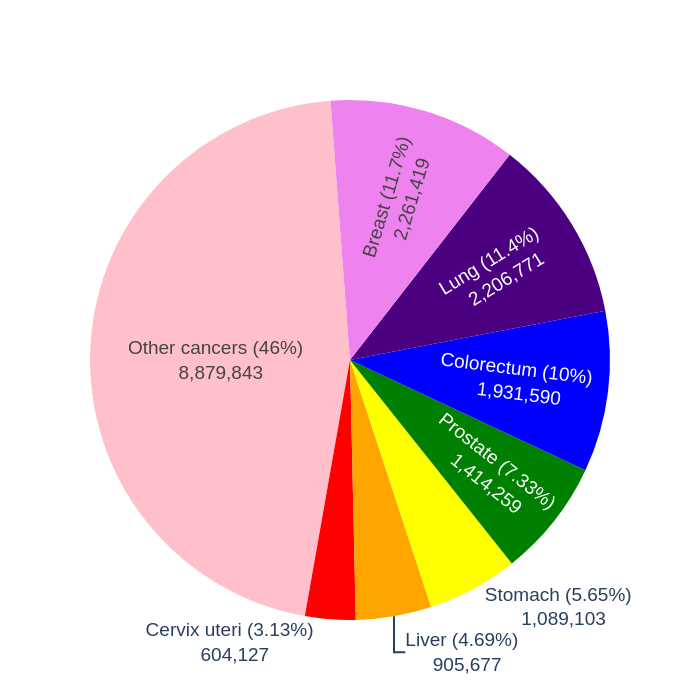}
    \caption{Estimated number of cancer incidences worldwide in 2020 (including both sexes and all ages). Breast cancer remains the leading cause of cancer, followed by lung, colorectum, prostate, stomach, liver, and esophagus cancer (GLOBOCAN 2020 statistics)~\cite{sung2021global}.}
    \label{fig:cancerstatistics1}
\end{figure}

\begin{figure}[!t]
    \centering
    \includegraphics[trim=2cm 0.5cm 2.5cm 1cm, clip, width=0.9\linewidth]{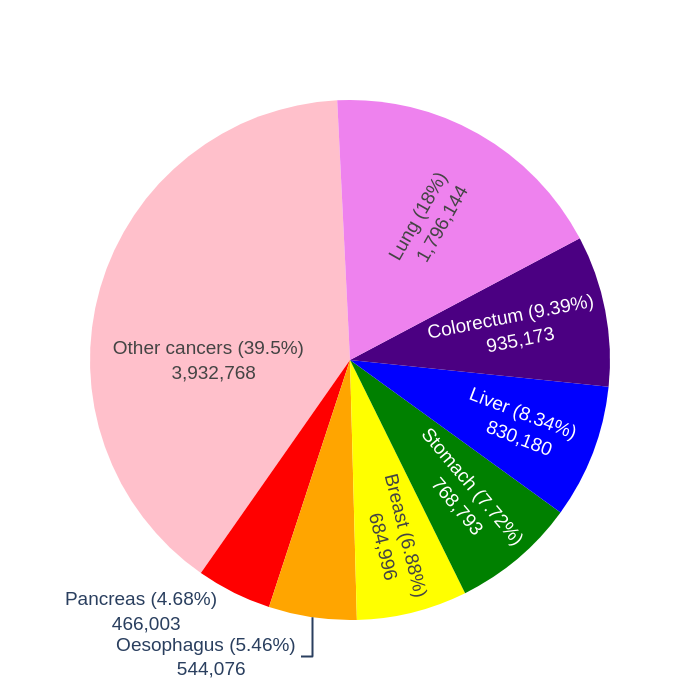}
    \caption{Estimated number of cancer mortality in 2020 worldwide, including both sexes and all ages. Lung cancer remained the leading cause of cancer-related mortality, followed by colorectum, liver, stomach, breast, oesophagus, pancreas, and other cancer (GLOBOCAN 2020 statistics)~\cite{sung2021global}.}
    \label{fig:cancerstatistics2}
\end{figure}

\vspace{1.0ex}

\noindent \textbf{Lung cancer detection}. Lung cancer is the second leading cause of cancer in terms of incidence and the leading cause of cancer-related death worldwide. It accounts for 11.4\% of all cancer-related incidence and accounts for 18\% of global cancer-related death~\cite{sung2021global}. Adnan et al.~\cite{adnan2022federated} studied the effects of IID and non-IID distributions along with the healthcare providers by conducting a case study. They proposed a deferentially private FL framework as the potential solution to analyze histopathology images. For the experimentation, they used the publicly available Cancer Genome Atlas (TCGA) dataset for simulating a distributed environment and empirically compared the performance of private, distributed training to conventional training. Their work revealed differential private learning as a potential framework for the collaborative and decentralized development of \ac{DL} models in medical image analysis tasks. In \cite{kennedy2022targeted}, a team of researchers from the University of Pennsylvania have used \ac{FL} to develop a model that can detect lung cancer in CT scans. The model was trained on data from 15 different hospitals, without any of the data ever leaving the individual hospitals. This allowed the researchers to build a model that is robust to the different imaging protocols used by different hospitals and researchers.

\vspace{1.0ex}

\noindent \textbf{Breast cancer detection}. Breast cancer is the first leading cause of cancer and the third leading cause of mortality. The Global Cancer Observatory (GLOBOCAN) accounts for 11.7\% of the incidence and 6.9\% of the mortality~\cite{sung2021global}. Roth et al.~\cite{roth2020federated} used FL to train a \ac{DL}-based model for breast density classification using a collaborative approach with seven clinical institutions worldwide. Their experimental results showed that the model trained under an FL setting outperformed the individually trained model on each institute's local data by 6.3\% on average. Additionally, a relative improvement of 45.8\% was observed in the model's generalizability when the FL-based model was evaluated on the external testing datasets from other participating sites. These improvements were explained by the larger training datasets obtained from seven different institutions under FL settings without requiring to share any data and respecting privacy. Similarly, Sanchez et al.~\cite{jimenez2021memory} defined a memory-aware curriculum learning method 
for the \ac{FL} settings. Their curriculum controlled the order of the training samples and paid special attention to forgotten samples after the deployment of the global model. Their approach combined the unsupervised domain adaptation to deal with the domain shift problem while preserving data privacy. They evaluated their method on three clinical datasets from different centers. Their results showed that federated adversarial learning was advantageous for multi-site breast cancer classification. In \cite{JIMENEZSANCHEZ2023107318}, the authors propose a novel method for \ac{FL} that uses curriculum learning to improve the classification performance of breast cancer models. Their method, called memory-aware curriculum \ac{FL}, prioritizes the training samples that are forgotten after the deployment of the global model. This helps to ensure that the local models are better able to generalize to new data. Other researchers from the University of Pennsylvania have implemented a \ac{FL}-based to calculate breast percent density~\cite{muthukrishnan2022mammodl}.

\vspace{1.0ex}
\noindent \textbf{Brain cancer detection}. \ac{FL} can be used to develop a model that performs brain cancer segmentation in medical imaging data, such as \ac{MRI} scans. The model can learn from diverse datasets from different hospitals, enhancing its accuracy and generalization across various brain cancer types~\cite{sheller2020federated}. Brain tumour segmentation is an important process for separating tumours from normal tissue. In the clinical setting, it can provide key details for each part of the tumour that is useful for diagnosis and treatment planning. \ac{MRI} has been extensively used by medical professionals for diagnosing brain tissue~\cite{bakas2017advancing}. However, a manual process is time-consuming and can only be accomplished by professional neuroradiologists~\cite{tang2018multi}. An automatic brain tumour segmentation can significantly impact brain tumour diagnosis and treatment. Li et al.~\cite{li2019privacy} implemented the feasibility of using the differential-privacy technique to protect patient data in the \ac{FL} setup. As a use case, they considered the practical learning systems for brain tumour segmentation on the BraTS 2018 dataset~\cite{bakas2017advancing,menze2014multimodal}. For the evaluation, they compared federated vs centralized data training. Their decentralized model converged at 300 training epochs, whereas the \ac{FL} converged at about 600 epochs. The data-centralized training took about 205.70s per epoch, whereas 65.45s plus small overheads.

Similarly, Sheller et al.~\cite{sheller2018multi} performed a multi-institutional study where they studied the feasibility of brain tumour segmentation. Their federated semantic segmentation model achieved a dice coefficient of 0.8520 on multimodel brain scans and achieved a dice coefficient of 0.8620 with the model trained by sharing the dataset. Due to the lack of a centralized public dataset, they used 2018 BraTs data for training their \ac{FL} based effective segmentation model by iteratively aggregating the locally trained UNet segmentation models at a centralized server. They compared the performance of \ac{FL} models with traditional institutional incremental learning (IIL) and cyclic institutional incremental learning (CILL) and showed improvement over both techniques.

}


\vspace{1.0ex}
\noindent \textbf{Colorectal polyp segmentation}.
Colorectal cancer is the third leading cause of cancer-related incidence and the second leading cause of death~\cite{sung2021global}. Colonoscopy is the gold standard for the early detection of cancer. However, there is a high adenoma miss-rate of 20-24\%~\cite{leufkens2012factors}. Moreover, the examination procedure is expensive and resource-demanding. A computer-aided diagnosis system can help detect neoplastic polyps during screening and aid in the early detection of cancer. FL can be one of the potential solutions to aid the development of computer-aided diagnosis systems for colonoscopy screening. To the best of our knowledge, there is no collaboration on developing decentralized AI in colorectal polyp segmentation. The reason for this is that the availability of a diverse and multi-center dataset remains one of the major challenges. To address the challenge of the lack of a multi-center dataset in the colonoscopy domain,  Ali et al.~\cite{ali2021polypgen} have curated and publicly released the PolypGen dataset, a six-center dataset that is suitable for exploring \ac{FL}. PolypGen consists of 3762 polyps frames from more than 300 patients' datasets. It consists of datasets from six unique clinical institutions (Norway, France, Italy, Egypt, and the UK). Each center dataset is stored in a separate folder. Thus, this dataset can be advantageous for training multiple \ac{DL}-based models independently on the colonoscopy datasets on separate computers and sharing the learned model weight. Although there has been work done on PolypGen dataset~\cite{ali2022assessing}, training the model on the FL-based paradigm still remains an open research area.

\vspace{1.0ex}
\noindent \textbf{Prostate cancer segmentation}. Prostate cancer is the second leading cause of cancer-related death (after lung cancer) in the American male population~\cite{aldoj2020semi}. An accurate and automatic prostate segmentation can play an important role in facilitating diagnostic and therapeutic applications such as early detection of cancer, patient management, treatment planning, and surgical planning~\cite{wang2015towards}. Manual segmentation is the commonly used method for the segmentation of prostate and prostate regions despite being a time-consuming and subjective process. In this regard, Sarma et al.~\cite{sarma2021federated} performed a multicenter study on the whole prostate segmentation on the axial T2- weighted \ac{MRI} scans. They used the 3D Hybrid Anisotropic Hybrid Network~\cite{liu20183d} model to train on each institution's data. Additionally, a  single model was trained using \ac{FL} across all three institutions. \ac{FL} based model had better performance and generalizability than the model trained on single institutions. 

Xia et al.~\cite{xia2021auto} proposed addressing the federated domain generalization problem. They proposed Federated Domain Generalization (FedG), a generalizable model that aims to learn a federated model from multiple distributed source generalizations that suffered from performance drops on unseen data outside of the federation. They trained FedG on a six prostrate dataset. The experimental results showed superior efficacy of FedG over the state-of-the-art methods. \textcolor{black}{Similarly, Liu et al.~\cite{liu2021feddg} highlighted a critical challenge in FL, where models trained within the federation may experience a significant performance drop when tested on entirely new hospitals located outside of the federation. In response to this challenge, they introduced a pioneering solution known as ``Episodic Learning in Continuous Frequency Space (EL-CFS)’’. This approach empowers each client to leverage multi-source data distributions, all while navigating the complex landscape of decentralized data. The effectiveness of EL-CFS becomes evident through its application to two distinct medical image segmentation datasets. The first dataset involves prostate MRI segmentation, encompassing data from six different medical centers. The second dataset comprises retinal fundus images, specifically focusing on the optic disc and optic cup, collected from four diverse medical centers. Comparative analyses, both qualitatively and quantitatively, consistently demonstrate the superior performance and efficiency of the EL-CFS algorithm.}

\vspace{1.0ex}
\noindent \textbf{Small bowel disease detection}.
Capsule endoscopy is the procedure to examine small intestine disease. The physicians recommend a capsule endoscopy procedure for the identification of gastrointestinal bleeding, diagnosing inflammatory disease (Crohn's disease), cancer, small intestine polyp, etc. The development of a \ac{DL}-based computer-aided diagnosis system can improve the anomaly detection rate in the small intestine and reduce manual labour. However, the capsule endoscopy dataset is scarce, and no multi-center dataset is publicly available for experimentation.  One of the large-scale datasets in this field is KvasirCapsule dataset~\cite{smedsrud2021kvasir} that contains  4,741,504 image frames.  This dataset was collected at Baerum hospital at Vestre Viken (VV) Hospital Trust in Norway. VV consists of four hospitals; therefore, this dataset can be treated as a multi-center dataset. However, the dataset from each center is not provided in a separate folder and no information is provided about the labelled and unlabelled videos at which hospital they were collected. To the best of our knowledge, there is no work on video capsule endoscopy for small bowel disease detection or gastrointestinal tract bleeding detection using \ac{FL}. Both exploring the dataset in the \ac{FL} setting and the use of \ac{FL} for small bowel disease detection remain open issues.

\vspace{1.0ex}
\noindent \textbf{Liver segmentation}.
Liver disease is of high importance because of its high incidence. It is the third leading cause of mortality in both men and women~\cite{sung2021global} (see Fig.~\ref{fig:cancerstatistics2}). An early diagnosis of liver disease can play an important role in treating and preventing damage. Liver segmentation is the most commonly adopted technique for diagnosis and follow-up treatment. \ac{FL}-based solutions recently gained popularity by training the shared global models by distributed clients with heterogeneous image datasets. Bernecker et al.~\cite{bernecker2022fednorm} proposed two \ac{FL} algorithms, namely FedNorm and FedNorm+, that were based on modality-based normalization techniques. They validated their method on the multi-modal and multi-institutional datasets (6 centers, 428 patients). They obtained the high performance of dice coefficients up to 0.9610 and consistently outperformed locally trained models that were trained in each center. In some cases, the federated models also even surpassed centralized models. Similarly, recently Xu et al.~\cite{xu2022federated} presented a new federated multi-organ segmentation U-Net for the segmentation of multiple organs. Here, the multi-encoding U-Net (MENU-Net) extracts organ-specific features via encoding sub-networks. Out of the different sub-networks, each of the sub-network is considered an expert of a specific organ and is trained for that client. They have regularized the training of the MENU-Net by designing an auxiliary generic decoder. This was performed to encourage organ-specific feature enhancement (for example, becoming distinctive and informative). They have trained their method on four abdominal CT image datasets (liver, kidney, pancreas, and BTCV). The extensive experimental results on four different datasets showed the potential of the proposed method in solving the partial level problem in \ac{FL} context.

\vspace{1.0ex}
\noindent \textbf{Pancreas segmentation}.
Pancreas cancer is among the most lethal malignant tumours characterized by high mortality~\cite{hidalgo2010pancreatic}. The GLOBOCAN 2020 statistic shows that it is the seventh leading lethal cancer (see Fig.~\ref{fig:cancerstatistics2}). The five-year survival rate of the patient with the disease is less than nine years~\cite{rawla2019epidemiology}. Therefore, accurate segmentation of pancreatic cancer can play an important role in clinical diagnosis and treatment. In this respect, Shen et al.~\cite{shen2021multi} investigated heterogeneous optimization methods with abdominal CT images from the pancreas and pancreatic tumours in \ac{FL} based settings. They utilized three publicly available annotated pancreas segmentation datasets (one consisting of pancreas and tumour, whereas the other consists of healthy pancreas cases) to model three different heterogeneous clients during \ac{FL} settings. They evaluated each method's performance in the \ac{FL} settings and compared their proposed global model's performance with \emph{FedAvg} and \emph{FedProx}. Their experimental results revealed that global models were better than local models for all datasets. Similarly, Wang et al.~\cite{wang2020automated} trained the \ac{FL} models for the segmentation of the pancreas without data sharing. Their dataset (abdominal CT images) consisted of healthy and unhealthy pancreas collected from two different institutions in Taiwan and Japan.  The model utilized for segmentation was coarse-to-fine network architecture search~\cite{yu2020c2fnas} along with variational auto-encoder~\cite{myronenko20183d} at the encoder endpoint. The quantitative results comparison between \ac{FL} settings and locally trained algorithms showed that \ac{FL} framework was better in dealing with unbalanced data distributions and successfully delivered better generalizable models compared to that of standalone training. 

\noindent \textbf{Treatment planning}.
{
\color{black}
Beyond its diagnostic applications, FL finds valuable utility in enhancing treatment responses. A prominent example of its application in the medical field lies within radiation therapy. FL-based models provide a collaborative solution that empowers radiation oncologists to devise the most optimal treatment plans tailored to individual patients, a process known as personalized treatment planning~\cite{smith2016personalized}. This approach proves particularly advantageous in conditions like lung cancer, where radiation therapy is utilized. The scope of FL's contributions extends beyond personalized treatment. It encompasses various facets of radiotherapy, including contouring, optimal dose prediction, tumor motion forecasting, image guidance, and survival prognosis. These advancements pave the way for precision and effectiveness in patient treatment, ultimately benefiting those undergoing therapy. Moreover, the incorporation of FL not only optimizes therapeutic outcomes but also ensures the security and confidentiality of patient data in alignment with regulatory standards. This dual focus on treatment efficacy and data privacy safeguards sensitive medical information. As distributed technology continues to evolve, we can anticipate further innovative applications of FL in the ongoing battle against cancer and other chronic diseases.}



\color{black}
\begin{table}
\centering
\caption{\textcolor{black}{Enhanced performance and generalizability of FL in clinical use~\cite{sarma2021federated}.}}
\label{table:FL_Gen}
\def\arraystretch{1.5}
\begin{tabular}{lcc}
\hline \textcolor{black}{Architecture} &\textcolor{black}{Institution} &\textcolor{black} {Dice coefficient} \\
\hline \textcolor{black}{Private model} & \textcolor{black}{NCI} & \textcolor{black}{0.872 $\pm$ 0.062} \\
\textcolor{black}{Private model} & \textcolor{black}{SUNY} & \textcolor{black}{0.838 $\pm$ 0.043} \\
\textcolor{black}{Private model} & \textcolor{black}{UCLA} & \textcolor{black}{0.812 $\pm$ 0.136} \\ 
\textcolor{black}{FL Model} & \textcolor{black}{Collaborative} & \textcolor{black}{0.889 $\pm$ 0.036} \\
\hline
\end{tabular}
\end{table}

In~\cite{sarma2021federated}, the efficacy of FL has been exemplified through a multi-institutional collaboration involving three esteemed academic establishments: the University of California, Los Angeles (UCLA), the State University of New York (SUNY) Upstate Medical University, and the National Cancer Institute (NCI) in the United States. This endeavour harnessed authentic clinical prostate imaging data to underscore the substantial capabilities of the FL framework, as medical imaging data for each of these intuitions was vital, and data privacy was the stringent requirement. Applied to the domain of medical image analysis, the study focused on the intricate task of whole prostate segmentation, which constitutes an initial stride in the accurate diagnosis of cancer through \ac{MRI} scans and in guiding fusion-based interventions. This research manifests the broader implications of FL's proficiency, with specific emphasis on its generalization capacity. The approach employed entailed the training and aggregation of models in a federated manner, thereby engendering a composite model equipped with predictive weights of a generalized nature, adaptable to the idiosyncrasies of each institutional dataset. Remarkably, this composite model exhibited good generalizability, as corroborated by its superior performance on both held-out test sets from the individual institutions and on external validation dataset.

An insightful evaluation of the findings, as elucidated in Table~\ref{table:FL_Gen}, underscores the intricacies of performance. The private models, while showcasing a diverse spectrum of competencies on the ProstateX dataset~\cite{armato2018prostatex} (as evidenced by Dice coefficient values ranging from 0.812 to 0.872), were collectively overshadowed by the \ac{FL} model. This latter model, exemplifying an aggregate mean Dice coefficient of 0.889, emerged as the definitive winner in performance.

Employing rigorous statistical analyses, the study reaffirmed the empirical superiority of the FL paradigm over the individual private model. This substantiates the effectiveness and practical viability of FL's capacity to transcend the limitations of isolated data silos and underscores its pivotal role in bolstering generalizability and predictive efficacy across diverse healthcare datasets.
\color{black}


\section{Open-Source FL Software Frameworks}
\label{open_source_FL_Software}


\textcolor{black}{In this section, we delve into the landscape of open-source \ac{FL} software frameworks, pivotal tools for rapid prototyping, experimentation, and validation of novel FL algorithms. Open-source FL software in healthcare ensures transparency and trust. Healthcare practitioners and institutions can inspect the code, verify security measures, and understand the algorithms being used, which is crucial for maintaining data privacy and complying with healthcare regulations. Open source also promotes innovation. Developers from diverse backgrounds can freely build upon existing FL solutions, tailoring them to address unique challenges in healthcare, such as disease diagnosis, drug discovery, or treatment personalization. This accelerates the development of FL applications in the medical field.} While some proprietary FL frameworks exist, such as IBM FL~\cite{ibmfl2020ibm}, NVIDIA CLARA~\cite{clara}, Substra~\cite{galtier2019substra} and Sherpa.AI FL framework~\cite{rodriguez2020federated} their lack of comprehensive documentation and intricate details can hinder the seamless exploration of new FL methodologies. Consequently, we focus here on elucidating the available open-source FL software frameworks. The ensuing sub-sections unravel the nuances of each framework, and a comprehensive comparison is distilled in Table\ref{table:federated_frameworks}.

\begin{table*}
\centering
\caption{Comparison of open-source FL software frameworks.}
\label{table:federated_frameworks}
\begin{tabular}{l|c|c|c|c|c|c|c|c|c}
\hline
\textbf{Features}                    & \textbf{FATE} & \textbf{TFF}                                        & \textbf{OpenFL} & \textbf{Fed-BioMed} & \textbf{PySyft}                                     & \textbf{FedML}                                      & \textbf{LEAF} & \textbf{PaddleFL} & \textbf{PrivacyFL} \\ \midrule
FL model with different datasets     & Yes           & Yes                                                 & Yes             & Yes                 & Yes                                                 & Yes                                                 & Yes           & Yes               & Yes                \\ \midrule 
FL attack simulator                  & No            & No                                                  & No              & No                  & No                                                  & No                                                  & No            & No                & Yes                \\ \midrule 
Documentation and tutorials details   & Partial       & Partial                                             & Yes             & Partial             & Yes                                                 & Partial                                             & Partial       & No                & No                 \\ \midrule 
Other library support                & No            & No                                                  & Yes             & Partial             & Yes                                                 & No                                                  & Yes           & No                & No                 \\ \midrule 
Data Partitioning                    & Yes           & Partial                                             & Yes             & Partial             & Yes                                                 & Yes                                                 & Partial       & Yes               & Partial            \\ \midrule 
Hardware support                     & CPUs          & \begin{tabular}[c]{@{}c@{}}GPUs\\ CPUs\end{tabular} & CPUs            & \begin{tabular}[c]{@{}c@{}}GPUs\\ CPUs\end{tabular}                & \begin{tabular}[c]{@{}c@{}}GPUs\\ CPUs\end{tabular} & \begin{tabular}[c]{@{}c@{}}GPUs\\ CPUs\end{tabular} & CPUs          & CPUs              & CPUs               \\ \midrule
FL specific for medical applications & No            & No                                                  & Yes             & Yes                 & No                                                  & No                                                  & No            & No                & No                 \\ \midrule
\end{tabular}
\end{table*}

\subsection{Federated AI Technology Enabler Framework (FATE)}
FATE, is a compelling open-source endeavor spearheaded by Webank, engineered to establish a secure computing platform for the federated AI ecosystem~\cite{fate_2019}. With FATE, diverse secure computation protocols seamlessly interlace, enabling collaborative data analysis while adhering to stringent data protection regulations. Notably, FATE’s script-based interface streamlines integration, albeit relying on command-line options and a specialized yet underspecified domain-specific language.

\subsection{\ac{TFF}}
\ac{TFF}, a prominent open-source venture, serves as a foundational framework for decentralized \ac{ML} on distributed data~\cite{tff}. TFF boasts dual layers—FL API and Federated Core API—catering to both ML developers and systems researchers. This multi-layered approach enables innovative algorithmic experiments, simulation of FL protocols, and model compatibility with TensorFlow models. However, it's noteworthy that TFF's arsenal lacks dedicated mechanisms for differential privacy, posing a minor constraint for privacy-preserving innovations.

\subsection{Open-\ac{FL} (OpenFL)}
OpenFL emerges as a versatile open-source framework specifically tailored for training \ac{ML} models under the paradigm of FL, particularly in the medical landscape~\cite{reina2021openfl}. This dynamic framework harmoniously coexists with TensorFlow and PyTorch training pipelines, and its adaptability extends to a plethora of ML and \ac{DL} frameworks. With OpenFL, practitioners can power the Federated Tumor Segmentation (FeTS) initiative (Federated Tumor Segmentation)\footnote{https://www.med.upenn.edu/cbica/fets/}, a real-world medical FL platform for tumour segmentation. This endorsement underscores OpenFL's suitability for medical applications, where data privacy and collaboration are paramount. The framework's architecture, encompassing both collaborator and aggregator components, orchestrates the training process on local datasets and the subsequent aggregation of model updates, facilitating collaborative learning. However, OpenFL currently lacks dedicated differential privacy mechanisms.

\subsection{Fed-BioMed}
Fed-BioMed\footnote{https://fedbiomed.gitlabpages.inria.fr/} is an open-source initiative aimed at enabling biomedical research using non-centralized statistical analysis and \ac{ML} methodologies~\cite{silva2020fed}. The project, which is currently based on Python, PyTorch, and Scikit-learn, allows for the development and deployment of FL analysis in real-world \ac{ML} systems. Fed-BioMed is an ongoing initiative, and it still needs to be developed to be scalable and fully deployed in a practical healthcare setting.

\subsection{PySyft}
PySyft is an open-source Python3 library that utilizes FL, differential privacy, and encrypted computations to enable FL for research purposes~\cite{pys}. It was created by the OpenMined\footnote{https://www.openmined.org/} community and mostly works with \ac{DL} frameworks like PyTorch and TensorFlow. PySyft defines objects, \ac{ML} methods, and abstractions. With PySyft, it is not possible as of now to work on ML/AI projects requiring network communication. This would necessitate the use of another package known as PyGrid\footnote{https://blog.openmined.org/what-is-pygrid-demo/}. Moreover, it does not provide any DP mechanism nor any DP algorithm.

\subsection{FedML}
FedML positions itself as a research-oriented open-source benchmark and framework, expertly engineered to streamline FL algorithm development and performance evaluations~\cite{he2020fedml}. This versatile framework accommodates diverse computing paradigms, ranging from on-device training for edge devices to distributed computing and single-machine simulations. Its lightweight Edge AI SDK finds application in a range of hardware setups, including edge GPUs, smartphones, and IoT devices. It's worth noting that while FedML’s potential is significant, its current documentation requires enhancement, and differential privacy components are not yet integrated.

\subsection{LEAF}
LEAF, a benchmarking framework dedicated to FL, serves as an invaluable resource for researchers probing diverse domains like multi-task learning, meta-learning, and on-device learning~\cite{caldas2018leaf}. Crafted through a mosaic of open-source datasets, statistics, and reference implementations, LEAF empowers researchers to explore novel methodologies with a level of pragmatism unprecedented in prior benchmarks. It’s pertinent to observe that while LEAF offers a substantial suite of resources, official documentation and dedicated benchmarks for differential privacy are areas that warrant further attention.

\subsection{PaddleFL}
PaddleFL, fortified by PaddlePaddle, emerges as a robust open-source FL framework designed to cater to industrial-scale parallel distributed \ac{DL}~\cite{paddle}. Its architecture equips researchers to swiftly reproduce and evaluate diverse FL techniques and algorithms, ensuring compatibility with an array of parallel distributed clusters. Most notably, PaddleFL leverages differential private stochastic gradient descent, offering enhanced privacy-preserving capabilities. It's essential to acknowledge that PaddleFL, despite its strengths, currently grapples with inadequately detailed documentation and a need for streamlined extensibility.

\subsection{PrivacyFL}
PrivacyFL, a scalable and configurable open-source FL framework, positions itself as a versatile solution that couples advanced analytics with security measures~\cite{mugunthan2020privacyfl}. Armed with features like latency simulation, resilience against client departure or failure, and differential privacy mechanisms, PrivacyFL caters to diverse requirements of FL within an array of scenarios. It's important to note that while PrivacyFL excels in configurability, it remains an emerging framework and could benefit from comprehensive documentation to ease the implementation of new FL algorithms.\\

\color{black}
Overall, these open-source software development for FL Learning in healthcare not only democratizes access to advanced machine learning tools but also encourages collaboration, innovation, transparency, and the growth of a vibrant community dedicated to improving patient care and healthcare outcomes. Based on our findings, we have identified OpenFL and Fed-BioMed as particularly suitable FL software frameworks for medical applications. OpenFL, with its emphasis on collaborative learning and compatibility with medical data paradigms, positions itself as a promising tool for enhancing medical research and diagnostics. On the other hand, Fed-BioMed's focus on enabling biomedical research through decentralized statistical analysis and \ac{ML} methodologies aligns well with the intricacies of medical data and privacy considerations.

\color{black}
\section{Challenges of \ac{FL} in Healthcare}
\label{opportunities_and_challenges_FL}

\textcolor{black}{
While FL offers significant promise in the context of the healthcare setting, it is essential to acknowledge and address its associated challenges and limitations. Some of the main challenges of FL in real-world medical environments include communication, statistical, computational, and security and privacy challenges~\cite{li2020federated}. In the following sub-sections, we explain these challenges in detail, emphasizing their relevance within the medical domain.}


\subsection{Communication Challenges}
Effective communication is important in \ac{FL}, where nodes or clients exchange information in a centralized FL system or within a distributed setup. FL requires multiple communication rounds among these entities to train a global model, which can strain communication bandwidth resources~\cite{zheng2020design}. In healthcare, where facilities may be geographically dispersed, unreliable or unstable network connections can hinder the FL process. This network reliability disparity across healthcare providers poses a significant hurdle, potentially impeding collaborative training of robust machine learning models. Here are some key aspects to consider:

\begin{itemize}
\vspace{1.0ex}
    \item \textbf{Bandwidth Bottlenecks}. \ac{FL} necessitates continuous communication between clients and a central server or between clients themselves in distributed FL. Each communication round involves the exchange of model updates or gradients, which can consume substantial communication bandwidth. In healthcare environments with a high number of participating nodes, this can strain available resources, leading to bandwidth bottlenecks that slow down the FL process.

    \vspace{1.0ex}
    \item \textbf{Geographical Disparities}. Healthcare providers are often geographically dispersed. This geographical diversity can result in unreliable or unstable network connections, which can hinder the smooth flow of data and model updates. Rural healthcare facilities, in particular, may face connectivity challenges, making it difficult for them to actively participate in FL processes.
    
    \vspace{1.0ex}
    \item \textbf{Latency Issues}. Real-time decision-making is critical in healthcare, and FL's reliance on communication rounds introduces latency. For time-sensitive applications, such as telemedicine or remote patient monitoring, delays in communication can have adverse consequences. Reducing latency while ensuring data security remains a complex challenge in implementing FL in healthcare.
    
    \vspace{1.0ex}
    \item \textbf{Data Privacy Concerns}. Communication during FL must be secure to protect sensitive healthcare data. Ensuring end-to-end encryption and data integrity during data transmission is crucial to mitigate privacy risks. Healthcare regulations, such as \ac{HIPAA} in the United States and \ac{GDPR} in Europe, impose stringent data security and privacy requirements, making robust communication protocols essential.
\end{itemize}


\subsection{Statistical Challenges}

The statistical challenge, i.e., data distribution among the available clients in the network participating in the FL process, is one of the main challenges of FL, particularly in healthcare applications. EHR is considered one of the prominent sources of healthcare data for FL applications. A detailed survey of existing works on EHR data for FL applications is outlined in~\cite{dang2022federated}. It is to be noted that healthcare data is inherently diverse and subject to variations across different facilities and regions. Here's a deeper exploration of these challenges:

\begin{itemize}
\vspace{1.0ex}
    \item  \textbf{Data Heterogeneity}. Healthcare data comes from various sources including EHRs, medical imaging, wearable devices, and genomics. Hence, handling and federating this heterogeneous data is understandably a significant challenge. EHRs are a primary data source for healthcare FL, but they can be heterogeneous due to variations in data formats, coding systems, and data collection practices. Integrating and harmonizing such diverse data sources for effective FL is challenging.

    \vspace{1.0ex}
    \item \textbf{Bias and Generalizability}. Training machine learning models on data from a single hospital or clinic can introduce bias and limit the model's generalizability. Patient populations, disease prevalence, and treatment protocols may vary across healthcare facilities, leading to model biases that may not translate well to other settings. Achieving model fairness and robustness against bias is a complex statistical problem.
    
    \vspace{1.0ex}
    \item \textbf{Data Sparsity}. Smaller healthcare clinics, especially in remote or underserved areas, may have limited EHR data. This data sparsity can hinder their participation in FL processes. Moreover, these clinics may lack the resources to label medical data comprehensively, leading to unlabelled data, which is challenging to leverage effectively in machine learning.

    \vspace{1.0ex}
    \item \textbf{Non-I.I.D. Data Distribution}. The common assumption in federated optimization is Identically and Independently Distributed (I.I.D.) data across clients. However, in healthcare FL, this assumption often doesn't hold true due to the inherent variations in healthcare data. Non-I.I.D. data distribution complicates the optimization process, increases the risk of straggler clients, and may add modelling, analysis, and evaluation complexity for FL algorithms~\cite{ding2022interval}.
\end{itemize}


\subsection{Computational Challenges}
It has been predicted that more than 80\% generated data by the edge devices will be processed at the edge of the network itself by 2030~\cite{peltonen20206g}. Thus, there is an increased interest in using AI either on edge devices or at the edge of the network. The computation challenges in healthcare-based FL are intertwined with the increasing adoption of Edge-AI and the diverse computational capabilities of healthcare devices. These challenges are pivotal for efficient model training and deployment:

\begin{itemize}
\vspace{1.0ex}
    \item \textbf{Edge Device Proliferation}. The healthcare landscape is witnessing a proliferation of edge devices, including wearable health gadgets, smartwatches, and IoT sensors. These devices generate a substantial volume of healthcare data and present opportunities for on-device AI. However, their computational capacities, storage capabilities, and energy profiles vary significantly.

    \vspace{1.0ex}
    \item  \textbf{Device Reliability}. Many edge devices are energy-constrained and may stop working during FL iterations. This can affect data quality and introduce unpredictability into the FL process. Ensuring that these devices reliably participate in FL while managing their limited resources remains a computational challenge~\cite{lo2022architectural}.

    \vspace{1.0ex}
    \item \textbf{Lightweight Models}. In cross-device FL settings, where on-device AI is prevalent, machine learning models must be lightweight and energy-efficient to run effectively on resource-constrained devices. Balancing model performance with resource constraints is an ongoing challenge in healthcare FL.

    \vspace{1.0ex}
    \item \textbf{Data Disparity in Cross-Silo FL}. Cross-silo FL, involving multiple healthcare institutions, is a common architecture in healthcare. However, the data disparity among these institutions may require a higher number of FL iterations to achieve the desired model accuracy on a global server~\cite{diddee2020crosspriv,nguyen2022novel}. This increased computation time poses challenges in maintaining efficiency and responsiveness in healthcare applications.

    \vspace{1.0ex}
    \item \textbf{Scalability}. \ac{FL} models can be computationally expensive to train and deploy. Therefore, scalability can be a critical challenge for modern healthcare organizations that have limited resources~\cite{li2020federated, diaz2023study}.
    
\end{itemize}


\subsection{Security and Privacy Challenges}
Security and privacy are paramount in FL-based healthcare due to the sensitive nature of medical data. Ensuring data protection while fostering collaborative model training is a complex endeavor. Moreover, the use of \ac{FL} in the healthcare setting is subject to a variety of regulations which can vary from country to country~\cite{fernandez2013security}. Hence, compliance with healthcare regulations like \ac{HIPAA}~\cite{wu2012towards, annas2003hipaa, choi2006challenges} and \ac{GDPR}~\cite{yuan2019policy} adds complexity to the security and privacy implementations in FL settings. As shown in Fig.~\ref{Federated_Learning_Architecture}, without having direct access to the client's local data, FL systems allow distributed clients to collaborate and train \ac{ML} and \ac{DL} models by sharing training parameters. However, instead of centralizing raw data, sharing gradient updates to a central server trained on the local clients could lead to reverse engineering attacks by passively intercepting the gradients exchanged during the training process~\cite{fredrikson2015model, fredrikson2014privacy}. This means malicious clients can introduce a backdoor functionality that compromises the underlying FL system during the training process of the global federated model~\cite{bagdasaryan2020backdoor}, which can be against the \ac{GDPR} since medical data are highly sensitive and private data. An attacker can use a backdoor functionality to mislabel specific jobs without affecting the overall accuracy of the global model. One possible option to address such a critical issue is to encrypt the global gradients shared in the distributed network. Differential privacy at the client level can help protect against backdoor attacks, but it comes at the cost of significantly reducing the performance of the global model~\cite{geyer2017differentially}. Other existing technical approaches also prevent a passive attacker from violating the privacy of data and information leakage by exploiting the global model outputs using malicious model updates. For example, the work in~\cite{jayaram2020mystiko} presents a novel cryptographic key generation and sharing approach that leverages additive homomorphic encryption to maximize the confidentiality of federated gradient descent in the training of deep neural networks without any loss of accuracy. The work in~\cite{mothukuri2021survey} identifies and examines some of the security vulnerabilities and threats in FL systems and provides insights into the existing defence techniques and future approaches for improving the security and privacy of FL implementations.

In summary, deploying FL in healthcare presents multifaceted challenges that demand innovative solutions addressing communication, statistical, computation, security, and privacy concerns. Addressing these challenges is essential to harness the full potential of FL while safeguarding the integrity and privacy of healthcare data.

\color{black}

\section{Open Problems and Future Research Directions of FL in Medical Context}
\label{Open_Problems_FL}

In the preceding sections, we discussed the evolving landscape of \ac{FL} within the healthcare domain and its profound potential. \textcolor{black}{However, FL in healthcare is still in its nascent stage, and numerous research endeavors are required to establish a practical framework for its application in medical settings. This section highlights the research gap and outlines promising avenues for future exploration in line with the earlier challenges.}

\subsection{{Efficient Hyperparameter Optimization}}
One of the open problems in the FL research is hyperparameter optimization. Even before starting an FL process, an AI model with properly optimized hyperparameters need to be designed that can be communicated from the FL server to multiple clients to train on their private medical data. This hyperparameter optimization is difficult to achieve on a federated medical dataset since the clients do not want to participate and contribute before benefiting from the FL algorithm through their involvement in the system. One of the possible ways to achieve this is to design an appropriate auction mechanism with incentives that would motivate the clients to contribute toward hyperparameter optimization~\cite{deng2021auction}.

\subsection{{Security and Privacy}}
\indent Security remains a paramount concern in FL systems, especially in medical contexts where privacy is a stringiest requirement. FL's collaborative nature, with participation from multiple clients, elevates susceptibility to security attacks, including model poisoning~\cite{bhagoji2019analyzing}. A malicious client can misclassify their input and influence the global model on the server. This situation is highly undesirable in a medical environment. Although some of the work in~\cite{malekzadeh2021dopamine} have introduced methods to distinguish between benign and malicious models. They have their own limitations, as such methods can only be applied in controlled settings under specific environments. Therefore, this is an open and challenging research issue that needs to be solved before deploying FL in a practical medical environment.

\subsection{{Efficient Communication Paradigm}}
\indent Communication efficiency is pivotal in FL systems, especially when numerous clients, such as smartwatches and health gadgets, are involved. To address the communication bottleneck, research must focus on designing lightweight AI models that maintain accuracy while minimizing the data transferred during uplink and downlink communications. Strategies such as sparsification, subsampling, and quantization offer avenues for reducing message sizes and optimizing bandwidth utilization~\cite{paragliola2022definition}. Additionally, exploring methods to minimize uplink communication rounds and enhance convergence speed while conserving client energy represents a promising research direction.

\subsection{{Solving Medical Data Heterogeneity and Statistical Issues}}
\indent Because of FL's distributed nature, data distributions across medical institutions are frequently heterogeneous. Several research studies have pointed out that the FL performance degrades with increasing degrees of data heterogeneity. Thus, another open problem to solve in FL is its medical data heterogeneity. FL can combine medical data, but combining horizontal medical data from medical institutions across regions and longitudinal medical data from the same patient across hospitals remains one of the most difficult challenges in FL. Although a few works such as~\cite{qu2021experimental,sheller2020federated} have investigated the medical data heterogeneity problem across multiple institutions, further research is essential to generalize results and address the nuances of medical data heterogeneity comprehensively.

\subsection{{Designing Intelligent Incentive Algorithms}}
\indent Motivating federated medical clients to actively participate in FL processes is vital for fostering the development of robust models. Crafting incentive mechanisms encouraging honest participation and rewarding high-quality data contributions is a pressing research area~\cite{yu2020fairness}. Simultaneously, developing intelligent FL algorithms that can discern and exclude clients with poor-quality data can enhance model generalizability~\cite{kang2019incentive}.

\textcolor{black}{\subsection{Integrating FL into the Foundational Models}}
\textcolor{black}{
\ac{FMs} are a large deep-learning network that is pre-trained on a massive amount of data and designed to be adopted for a wide variety of downstream tasks. FL can potentially empower the \ac{FMs}, encompassing \textit{pre-training, fine-tuning and downstream applications}. Through FL, it is possible to collect a large dataset from different medical centers across the border during the \textit{pre-training}. Collaborative learning can help develop robust and generalized FM models for various applications and domains. Additionally, FL can play an important role in \textit{fine-tuning} of FMs by incorporating incremental data, which helps to remain up-to-date and adaptable. One of the examples of such models is GPT-JT (with six billion parameters), which has outperformed many FMs with significantly larger parameter counts\footnote{\url{https://together.ai/blog/releasing-v1-of-gpt-jt-powered-by-open-source-ai}}. This shows that further \textit{fine-tuning} can improve the performance of FMs. Moreover, FL enables the distributed utilization of FMs from \textit{downstream tasks}. The distributed approach empowers diverse participants to collaborate, share insights, and drive continuous improvement in FM applications. Some examples of \textit{downstream tasks} include sentiment analysis, question answering, translation, image classification, object detection, speech recognition, etc. The downstream tasks benefit in terms of privacy preservation, better generalization, adaptability, and resource utilization.} 

\textcolor{black}{
\subsection{Generative Pre-trained Large Language Model (FL-GPT)}
The pre-trained large language models (LLM) mostly built upon the transformer architecture (e.g., BERT, GPT, Amazon Titan, AI21 Jurassic, Cohere), have garnered huge interest in healthcare applications, especially in administrative tasks (\textit{such as generating letters, discharge summaries in clinics and aiding in disease diagnosis. The other application includes helping medical professionals to use LLM's model (for example, ChatGPT) for training, education and clinical research}. However, the need for substantial healthcare data to effectively train these models raises privacy and security concerns. Addressing this, FL can be used as a crucial approach, offering privacy-preserving model training across decentralized healthcare institutions. FL not only safeguards sensitive patient information but also enables the development of specialized, context-aware models to suit the diverse healthcare landscape. The research direction calls for optimizing FL methodologies to ensure efficient, secure, and collaborative training of a specialized FL-enabled generative pre-trained large language model (FL-GPT) for healthcare, addressing communication efficiency, model convergence, and privacy preservation while enhancing collaboration and data integration among healthcare organizations. This trajectory holds great promise to redefine healthcare, ensuring patient data privacy while enhancing care and diagnostics.}


\textcolor{black}{
\subsection{Image Generative AI models}
Image generative AI models (for example, Stable Diffusion, DALL-E2, CLIP, and Imagen) help to generate realistic medical images for radiological applications. One example is that by using a stable diffusion model, it is possible to generate liver tumours with different sizes, shapes, orientations, textures, intensities, and locations. Synthetically generating hard-to-find tumours, such as small liver tumours with less than 10 mm in size or rare tumours, helps to include such datasets in the training process, which can further boost the network's performance, leading to better robustness and generalizability. Further optimization under the FL can improve the performance of different radiological imaging applications, such as diagnostic and prognosis assistance, helping in early disease diagnosis and treatment decisions.}


\subsection{{Contrastive Learning for Unlabelled Medical Data }}
\indent Most of the research works in FL are concentrated on the supervised learning models where the datasets are labeled~\cite{ji2021emerging,fallah2020personalized}. However, unlabelled datasets could grow in numbers when small health organizations and big hospitals participate in the FL for training a global model. To overcome this unlabelled datasets problem, semi-supervised and unsupervised learning have recently been used in FL by some researchers~\cite{kassem2022federated,itahara2020distillation,gornitz2014learning,zhu2021federated}. Although semi-supervised and unsupervised learning can solve the problem related to the unlabelled datasets, they often fail to achieve high performance in medical image datasets with \ac{DL} models. In this regard, contrastive learning has recently become a popular research field for learning unlabeled data representations by training an FL model on unlabeled data~\cite{li2021model}. Contrastive learning, also known as self-learning, is a pre-training procedure in which the model attempts to learn similar and distinct data samples from an unlabeled data distribution~\cite{he2020momentum,chen2020simple}. Thus, contrastive learning seems very relevant for applying FL in the medical domain as it can be used on unlabelled and non-IID medical data. This is an interesting research direction, and more research is needed to be explored in FL settings.

\subsection{{Benchmarking FL}}
\indent As FL gains prominence in the medical domain, it becomes imperative to establish robust benchmarking tools and frameworks for empirical evaluation and comparison. Expanding existing implementations and fostering the availability of open FL-based medical datasets for research communities are essential steps to promote reproducibility, generalizability, and innovation in FL solutions. Also, open FL-based medical datasets should be made available for the research community for benchmarking FL algorithms suitable for medical applications.

\subsection{{Regulatory Frameworks for Medical Applications}}
Considering the strict regulations such as GDPR on the security and privacy of medical data, a guideline needs to be developed that allows researchers to evaluate actual risks and concerns related to the medical data when using it in the FL framework. So far, we do not have enough knowledge and the possibility to check if privacy regulations of medical data are being violated when using it for FL applications. This calls for joint cooperation between computer science and law community researchers to address the security and privacy regulations of medical data in the FL framework. Moreover, enhancing the explainability of FL models in medical contexts is an open challenge that warrants exploration~\cite{samek2017explainable,chen2022evfl,wang2019interpret}.

\color{black}
\section{Conclusions}
\label{conclusion}
In this paper, we explored the pivotal role of FL in privacy-preserving medical applications where direct access to sensitive medical data is limited. \textcolor{black}{FL, being a decentralized and robust framework, is well-suited for healthcare. Our discussion has highlighted how FL and emerging technologies address the unique challenges in the medical field, especially in the context of addressing global cancer burdens. Our research demonstrates how FL can enable the creation of computer-aided diagnosis tools that are more effective than traditional data-driven medical applications. However, several challenges remain, such as system and statistical differences, communication limitations, and security and privacy concerns. We have given a comprehensive overview of FL and its applications, using cutting-edge technologies to overcome the intricate challenges in the medical domain. Throughout this article, we have presented the latest research findings showcasing the evolving use of FL in medical contexts. These findings emphasize the dynamic nature of FL research in healthcare. To inspire future research, we have explored specific challenges in FL in medical scenarios, including differences between healthcare institutions, communication issues in FL networks, and the need for strong security and privacy measures. These challenges define the current state of FL in healthcare and guide future investigations.} Addressing these challenges and advancing FL in medical applications requires collaboration across diverse interdisciplinary research communities, including health, computer science, and law. Combining expertise from these fields is crucial for designing reliable and scalable FL models that can drive healthcare innovations based on data.\\
\indent In future, we anticipate the practical deployment of FL for medical applications, leading to the development of more secured protocols. Additionally, we anticipate the emergence of energy-efficient communication paradigms tailored to the unique requirements of FL networks. The translation of FL technology into real-world healthcare solutions is poised to revolutionize the healthcare landscape by addressing complex challenges.\\
\indent In conclusion, this survey paper serves as a crucial reference, summarizing the current state of FL in medical applications. It also acts as a guide for researchers, pointing out open problems and future research directions in this field. We strongly believe that FL has the potential to greatly benefit healthcare by improving the efficiency of healthcare systems worldwide.

\bibliographystyle{IEEEtran}
\bibliography{Bibliography.bib}

\begin{IEEEbiography}[{\includegraphics[width=1in,height=1.25in,clip,keepaspectratio]{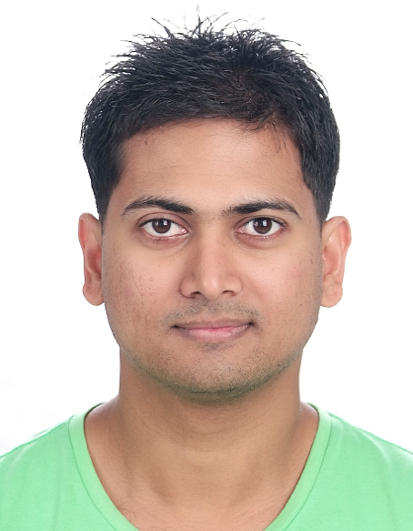}}]{ASHISH RAUNIYAR}
is currently working as a Research Scientist at SINTEF Digital, Norway.
He received a Ph.D. Degree in Computer Science from the University of Oslo, Norway in 2021. He was a graduate research assistant at Wireless Emerging Networking System (WENS) Lab, where he completed his Master's degree in IT Convergence Engineering at Kumoh National Institute of Technology, South Korea. He is a recipient of Best Paper Awards at the 2020 IEEE 43rd International Conference on Telecommunications and Signal Processing (TSP), Milan, Italy, 28th IEEE International Telecommunication Networks and Applications Conference (ITNAC), 2018, Sydney, Australia, and AI-DLDA 2018 International Summer School on Artificial Intelligence, Udine-Italy, 2018. He was also selected as ``Top 200 Young Researchers in Computer Science \& Mathematics'' and invited to attend Heidelberg Laureate Forum, Heidelberg, Germany in 2017, and Global Young Scientist Summit, Singapore in 2020. He also won the European Satellite Navigation Competition (ESNC) in 2017. His main research interest includes 5G/6G Signal Processing, Autonomous Systems and Networks, Internet of Things, Machine Learning, Wireless Communications, and Computer Networking.
\end{IEEEbiography}

\begin{IEEEbiography}[{\includegraphics[width=1in,height=1.25in,clip,keepaspectratio]{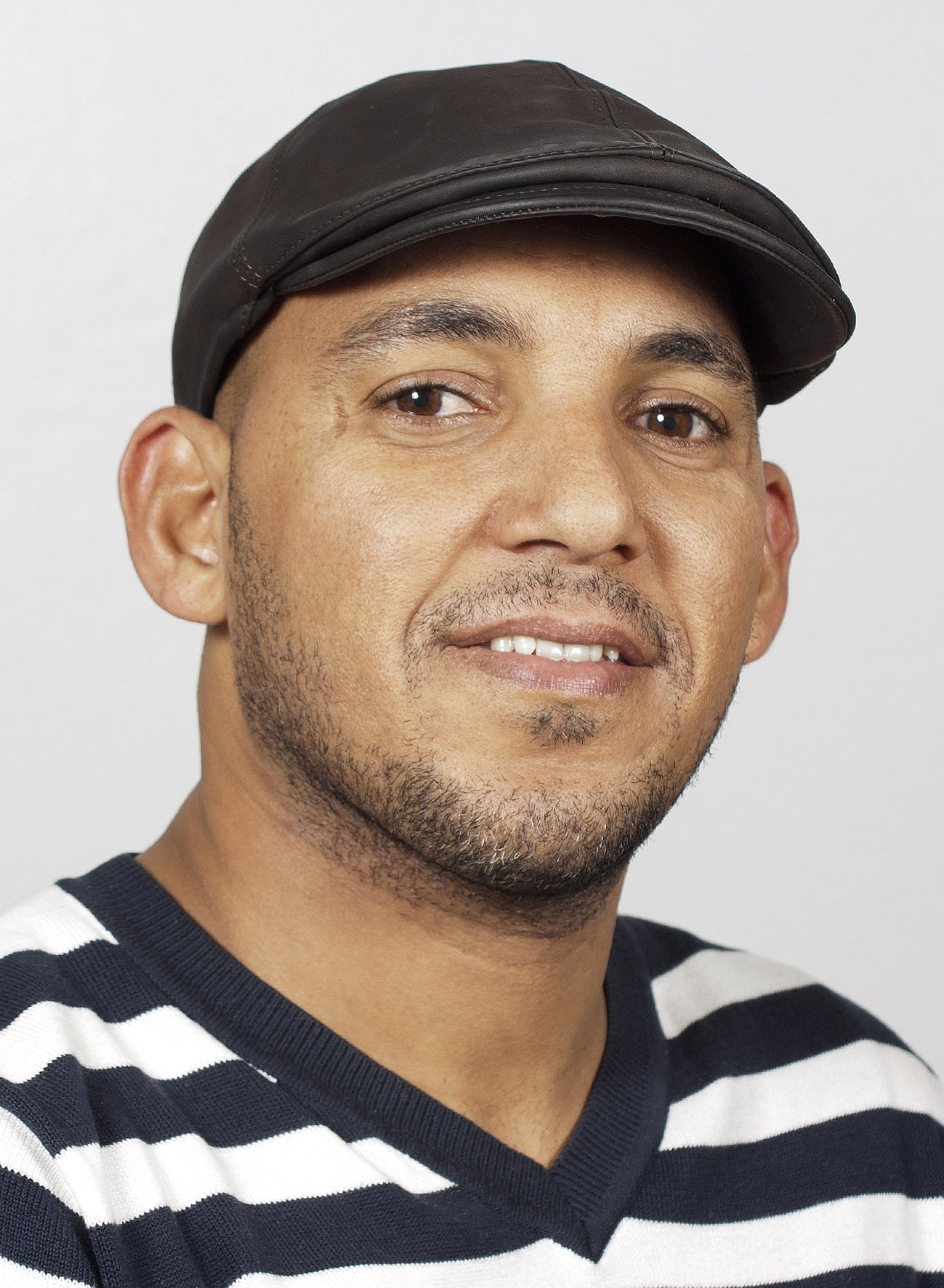}}]{DESTA HAILESELASSIE HAGOS}
received a Ph.D. degree in Computer Science from the University of Oslo, Faculty of Mathematics and Natural Sciences, Norway, in April 2020. Currently, he is a Postdoctoral Research Fellow at the DoD Center of Excellence in Artificial Intelligence and Machine Learning (CoE-AIML), College of Engineering and Architecture (CEA), Department of Electrical Engineering and Computer Science at Howard University, Washington DC, USA. Previously, he was a Postdoctoral Research Fellow at the Division of Software and Computer Systems (SCS), Department of Computer Science, School of Electrical Engineering and Computer Science (EECS), KTH Royal Institute of Technology, Stockholm, Sweden, working on the H2020-EU project, ExtremeEarth: From Copernicus Big Data to Extreme Earth Analytics. He received his B.Sc. degree in Computer Science from Mekelle University, Department of Computer Science, Mekelle, Tigray, in 2008. He obtained his M.Sc. degree in Computer Science and Engineering specializing in Mobile Systems from Lule\aa ~University of Technology, Department of Computer Science Electrical and Space Engineering, Sweden, in June 2012. His current research interests are in the areas of Machine Learning, Deep Learning, and Artificial Intelligence.
\end{IEEEbiography}

\begin{IEEEbiography}[{\includegraphics[width=1in,height=1.25in,clip,keepaspectratio]{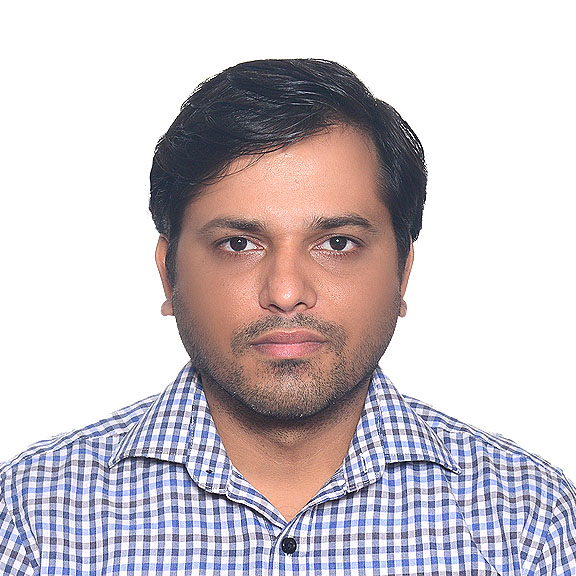}}]{DEBESH JHA} is currently working as a Research Associate at the Machine \& Hybrid Intelligence Lab, Department of Radiology, Northwestern University. He received a Ph.D. in Computer Science from the UiT The Arctic University of Norway in 2022. Previously, he worked as a researcher at UiT The Arctic University of Norway, Troms{\o}, Norway. During his Ph.D., he was also affiliated with Simula Research Laboratory, Oslo, Norway, and Simula Metropolitan Center for Digital Engineering, Oslo, Norway. He was a graduate research assistant at the Digital Media Computing (DMC) Lab, where he completed his Master's Degree in Information and Communication Engineering from the Chosun University, Republic of Korea. He is a recipient of the First-ever Paper with Code Contributor Award in 2022. He is also the recipient of the Best Paper Award at the International Conference on Electronics, Information, and Communication (ICEIC 2018), Hawaii, USA, and the Best Student Paper Award finalist at the IEEE 33rd International Symposium on Computer Based Medical Systems (CBMS 2020). His research interest includes Medical Image Analysis, Computer-aided Diagnosis and Detection, Machine Learning, Deep Learning, Computer Vision, and their application in Biomedical and Clinical Imaging. 
\end{IEEEbiography}

\begin{IEEEbiography}[{\includegraphics[width=1.10in,height=1.75in,clip,keepaspectratio]{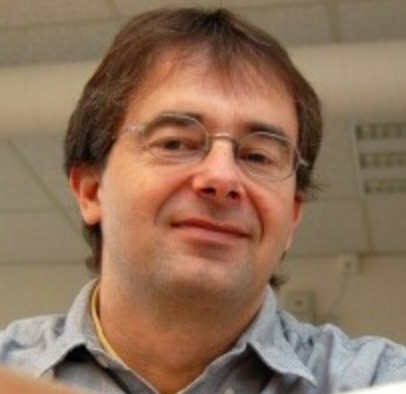}}]{JAN ERIK HÅKEGÅRD}
received the degree Sivilingeniør (M.Sc.) in 1990 from the Department of Electronical Engineering and Informatics, The Norwegian Institute of Technology (NTH), Trondheim, Norway. In 1997, he received a Docteur (Ph.D.) degree in Electronics and Communications at ENST, site de Toulouse, France. Since 1997, he has been with SINTEF Digital, working on research and development projects related to various types of Wireless Communication Systems. He is currently leading SINTEF activities within the Satellite and Terrestrial Communication Systems.
\end{IEEEbiography}
\vspace{-5ex}
\begin{IEEEbiography}[{\includegraphics[width=1.10in,height=1.75in,clip,keepaspectratio]{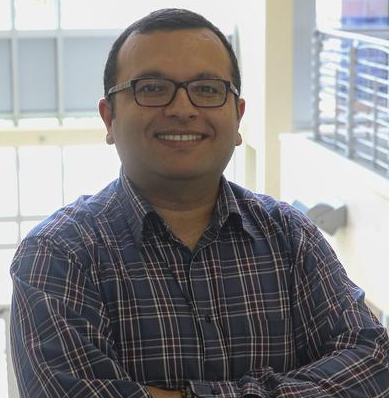}}]{ULAS BAGCI, Ph.D.,} is an Associate Professor at the Northwestern University's Radiology, BME, and ECE departments, and He holds a courtesy professorship at the Center for Research in Computer Vision (CRCV), University of Central Florida (UCF). His research interests are Artificial Intelligence, Machine Learning, and their applications in Biomedical and Clinical Imaging. Dr. Bagci has more than 250 peer-reviewed articles on these topics. Previously, he was a staff scientist and lab co-manager at the National Institutes of Health’s radiology and imaging sciences department, center for infectious disease imaging. Dr. Bagci holds NIH grants and serves as a steering committee member of AIR (artificial intelligence resource) at the NIH. Dr. Bagci also serves as an area chair for MICCAI for several years and he is an associate editor of top-tier journals in his fields such as IEEE Transactions on Medical Imaging, Medical Physics, and Medical Image Analysis. He has several international and national recognitions including Best Paper and Reviewer Awards. 
\end{IEEEbiography}

\begin{IEEEbiography}[{\includegraphics[width=1in,height=1.25in,clip,keepaspectratio]{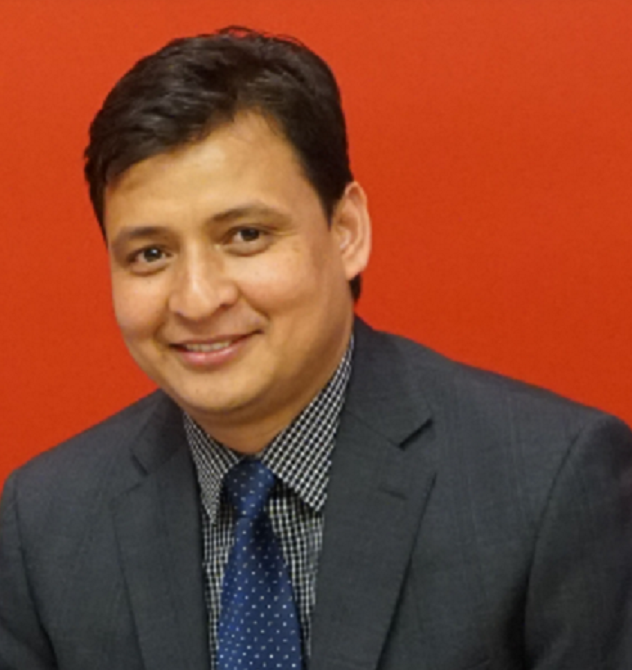}}]{Dr. DANDA B. RAWAT}
is the Associate Dean for Research \& Graduate Studies, a Full Professor in the Department of Electrical Engineering \& Computer Science (EECS), Founding Director of the Howard University Data Science \& Cybersecurity Center, Founding Director of the DoD Center of Excellence in Artificial Intelligence \& Machine Learning (CoE-AIML), Director of Cyber-security and Wireless Networking Innovations (CWiNs) Research Lab, Graduate Program Director of Howard CS Graduate Programs and Director of Graduate Cybersecurity Certificate Program at Howard University, Washington, DC, USA. Dr. Rawat is engaged in research and teaching in the areas of cybersecurity, machine learning, big data analytics, and wireless networking for emerging networked systems including cyber-physical systems (eHealth, energy, transportation), Internet-of-Things, multi-domain operations, smart cities, software-defined systems, and vehicular networks. 
\end{IEEEbiography}

\begin{IEEEbiography}[{\includegraphics[width=1in,height=1.25in,clip,keepaspectratio]{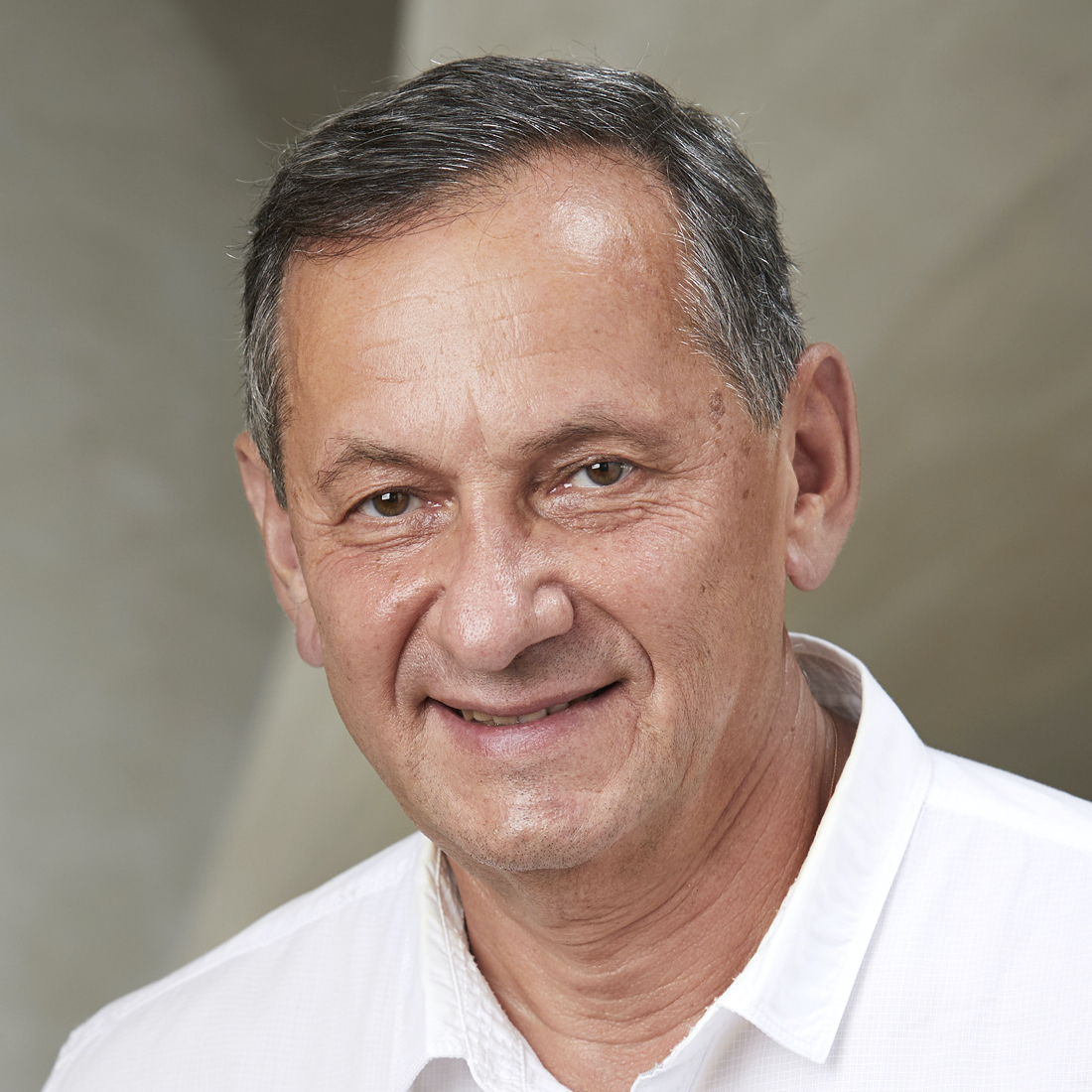}}]{VLADIMIR VLASSOV}
is a professor in Computer Systems at the Department of Computer Science, School of Electrical Engineering and Computer Science, KTH Royal Institute of Technology in Stockholm, Sweden. He was a visiting scientist with the Massachusetts Institute of Technology (1998) and at the University of Massachusetts Amherst (2004), USA. He has participated in a number of research projects funded by the European Commission, Swedish funding agencies, and the National Science Foundation (NSF) USA. He was one of the coordinators of the EMJD-DC Erasmus Mundus Joint Doctorate in Distributed Computing. Currently, he is a principal investigator from KTH in the H2020-EU project ``ExtremeEarth: From Copernicus Big Data to Extreme Earth Analytics” (2018-2020). His research covers several areas in computer science, including Data-intensive Computing, Stream Processing, Scalable Distributed Deep Learning, Distributed Systems, Autonomic Computing, Cloud, and Edge Computing.

\end{IEEEbiography}
\end{document}